\documentclass[sigconf]{acmart}
\usepackage{multirow}
\usepackage{caption,subcaption}
\usepackage{subcaption}
\usepackage{wrapfig}
\usepackage{graphicx}
\usepackage{bm}
\usepackage{algorithm}
\usepackage{algpseudocode}
\usepackage{placeins}
\AtBeginDocument{%
  }

\copyrightyear{2025}
\acmYear{2025}
\setcopyright{cc}
\setcctype{by}
\acmConference[CIKM '25]{Proceedings of the 34th ACM International
Conference on Information and Knowledge Management}{November 10--14,
2025}{Seoul, Republic of Korea}
\acmBooktitle{Proceedings of the 34th ACM International Conference on
Information and Knowledge Management (CIKM '25), November 10--14, 2025,
Seoul, Republic of Korea}
\acmDOI{10.1145/3746252.3761123}
\acmISBN{979-8-4007-2040-6/2025/11}

\newcommand{\model}{\textsc{EASE}}
\usepackage{fancyhdr}
\fancypagestyle{firstpage}{
  \fancyhead{} 
  \fancyhead[C]{\textit{Accepted at ACM International Conference on Information and Knowledge Management 2025 (CIKM'25).}}
  }

\begin{document}

\title{Adaptive Context-Infused Performance Evaluator for Iterative Feature Space Optimization}

\author{Yanping Wu}
\authornote{Work done during internship at University of Kansas.}
\affiliation{
  \institution{University of Kansas}
  \city{Lawrence}
  \country{USA}}
\email{yanping@ku.edu}

\author{Yanyong Huang}
\affiliation{
  \institution{Southwestern University of Finance and Economics}
  \city{Chengdu}
  \country{China}}
\email{huangyy@swufe.edu.cn}

\author{Zijun Yao}
\affiliation{
  \institution{University of Kansas}
  \city{Lawrence}
  \country{USA}}
\email{zyao@ku.edu}

\author{Yanjie Fu}
\affiliation{
  \institution{Arizona State University}
  \city{Tempe}
  \country{USA}}
\email{yanjie.fu@asu.edu}

\author{Kunpeng Liu}
\affiliation{
  \institution{Portland State University}
  \city{Portland}
  \country{USA}}
\email{kunpeng@pdx.edu}

\author{Xiao Luo}
\affiliation{
  \institution{University of California}
  \city{Los Angeles}
  \country{USA}}
\email{xiaoluo@cs.ucla.edu}

\author{Dongjie Wang}
\authornote{Dongjie Wang is the corresponding author.}
\affiliation{
  \institution{University of Kansas}
  \city{Lawrence}
  \country{USA}}
\email{wangdongjie@ku.edu}


\renewcommand{\shortauthors}{Yanping Wu et al.}

\begin{abstract}
 Iterative feature space optimization includes continuously evaluating and refining the feature space to improve downstream task performance.
 However, existing methods commonly suffer from three major limitations: 1) ignoring differences between samples leads to evaluation bias; 2) the feature space is overly tailored to specific models, resulting in overfitting and poor generalization; and 3) retraining the evaluator from scratch in each iteration significantly reduces overall efficiency.
    To bridge these gaps, we introduce \model\ (g\underline{\textbf{E}}neralized \underline{\textbf{A}}daptive feature \underline{\textbf{S}}pace \underline{\textbf{E}}valuator), a generalized framework for efficient and objective evaluation of iteratively generated feature spaces.
    This framework includes two key components: Feature-Sample Subspace Generator and Contextual Attention Evaluator.
    The first component aims to mitigate evaluation bias by decoupling the information distribution within the feature space.
    To achieve this, based on feedback from the subsequent evaluator, we identify the samples most challenging for evaluation and the features most relevant to prediction tasks.
    The second component intends to incrementally capture evolving patterns of the feature space for efficient evaluation.
    Specifically, we propose a weighted-sharing multi-head attention mechanism to encode the feature space into an embedding vector for evaluation, and update the evaluator incrementally to retain prior knowledge while incorporating new information.
    Extensive experiments on fifteen public datasets demonstrate the effectiveness of \model.
    We have released our code and data to the public~\footnote{\scriptsize\url{https://anonymous.4open.science/r/EASE-1C51}}.
\end{abstract}




\maketitle
\thispagestyle{firstpage}

\section{Introduction}
\begin{figure}[t]
\label{fig:feature_space}
    \subcaptionbox{Traditional feature space optimization framework.\label{fig:feature_space_a}}{
        \includegraphics[width=8cm,height=3.2cm]{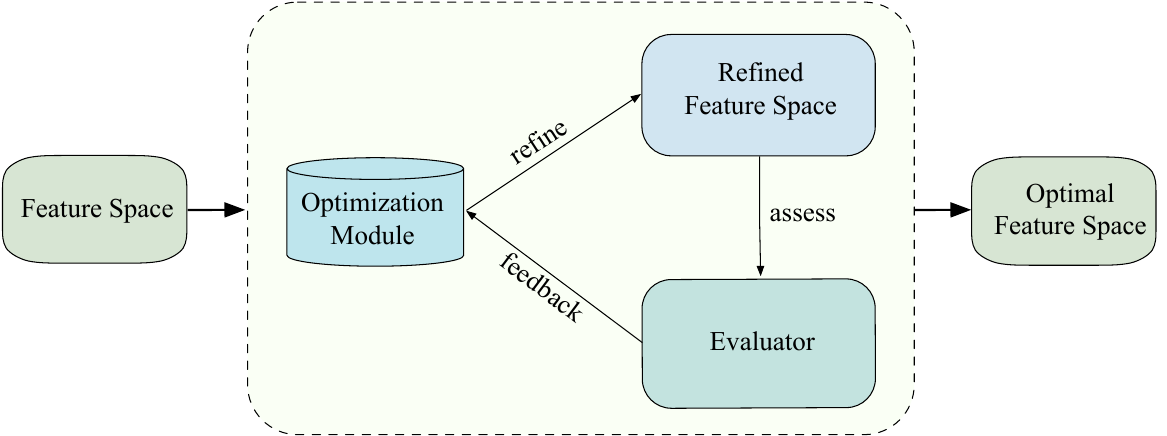}}\hfill
    \subcaptionbox{Our proposed feature space optimization framework.\label{fig:feature_space_b}}{
        \includegraphics[width=8cm,height=3.3cm]{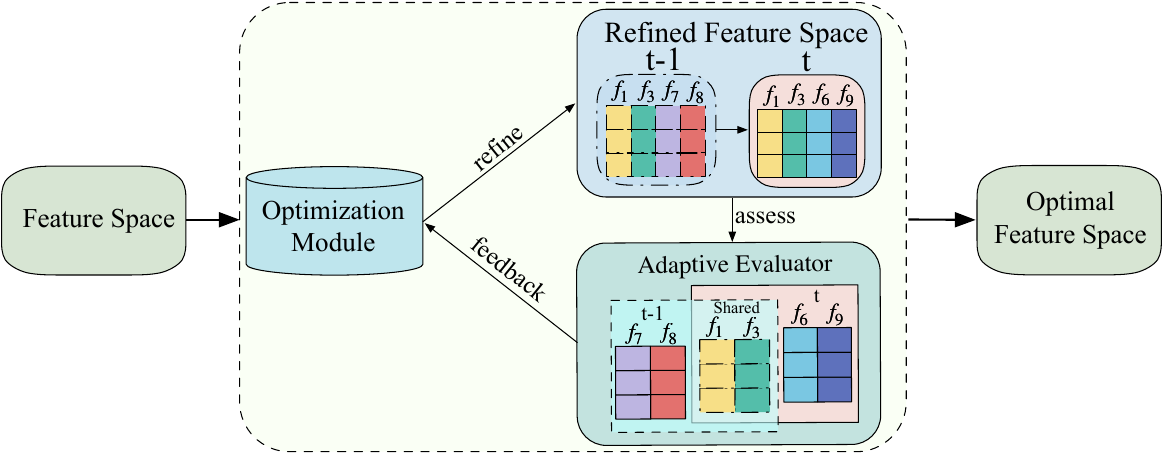}}  
    \caption{(a) Illustration of the iterative feature space optimization, where the optimization module refines the feature space based on the feedback of the evaluator until the optimal one is identified. (b) The feature spaces between consecutive iterations exhibit informational overlap.} 
\end{figure}

Iterative feature space optimization systematically evaluates and refines the feature space to enhance downstream task performance \citep{jia2022feature}. As depicted in Figure~\ref{fig:feature_space_a}, the optimization module iteratively enhances the feature space based on the feedback from the evaluator.
This optimization process continues until the optimal feature space is identified. 
This approach has demonstrated broad applicability and has been successfully adopted in various fields, including biology, finance, and medicine \citep{zhu2023hybrid,htun2023survey,vommi2023hybrid}.

Research in this domain has received significant attention \citep{zebari2020comprehensive}. Recursive optimization methods focus on evaluating feature importance to progressively refine the feature space \citep{darst2018using,priyatno2024systematic}. For instance, 
\citet{8614039} utilized sensitivity testing with membership queries on trained models to recursively identify key features. To improve the efficiency of these recursive methods, evolutionary algorithms and reinforcement learning (RL) were subsequently introduced, further accelerating the refinement process 
\citep{xiao2024traceable,wang2024mel,liu2021automated}. For example, \citet{wang2022group} employed three cascading agents to replicate the feature engineering process typically performed by human experts, using RL to streamline the exploration phase.

But, existing approaches suffer from three key limitations:
 \textbf{Limitation 1: Evaluation bias.} These methods do not account for variability between samples, which limits the evaluator's ability to capture the full range of features of the space. As a result, the performance assessments become biased and do not objectively reflect the quality of the feature space.
\textbf{Limitation 2: Non-generalizability.} Customizing the feature space by replacing the evaluator based on specific requirements limits its ability to capture generalizable patterns. Consequently, the resulting feature space lacks flexibility and cannot be effectively applied across diverse scenarios.
\textbf{Limitation 3: Training Inefficiency.} Retraining the evaluator from scratch at each iteration significantly increases computational demands. This technical trait leads to a time-consuming process that hinders efficiency and scalability.

Thus, there is a vital need for a robust evaluation framework that can efficiently assess feature space quality, enabling the creation of generalized and optimal feature spaces. This framework should integrate seamlessly with iterative feature space optimization algorithms to enhance their performance and efficiency. However, to accomplish this, there are two key technical challenges: 
\begin{itemize}
    \item \textbf{Challenge 1: Complicated Feature Interactions.} 
    Within the feature space, complex feature-feature interactions exist, which are important for understanding its characteristics and enabling more effective refinement.
    But, how can we effectively capture such complicated information as the guidance during the iterative optimization process?

    \item \textbf{Challenge 2: Incremental Evaluator Updates.}
    As illustrated in Figure~\ref{fig:feature_space_b}, the feature spaces between consecutive iterations exhibit partial overlap. This overlap presents an opportunity to update the evaluator efficiently, rather than retraining it from scratch.
    But, how can we incrementally update the parameters of the evaluator to ensure it retains essential prior evaluation knowledge while simultaneously integrating new evaluation insights?
\end{itemize}

To address these challenges, we propose \textbf{\model}, a g\underline{\textbf{E}}neralized \underline{\textbf{A}}daptive feature \underline{\textbf{S}}pace \underline{\textbf{E}}valuator, which can seamlessly integrate as a plugin into any iterative feature space optimization method for enhanced feature space refinement.
This framework contains two key components: Feature-Sample Subspace Generator and Contextual Attention Evaluator.
The first component aims to decouple the information distribution within the feature space to mitigate evaluation bias. 
To achieve this, we initially employ the feature index optimizer to select the features most relevant to the prediction task. 
Next, we use the sample index optimizer to identify the samples that present the greatest evaluation challenges.
Both the previous two steps were guided by feedback from the subsequent evaluator.
Finally, we use the identified features and samples to construct feature subspaces for the next iteration of feature space refinement.
This decoupling strategy enables the evaluator to consistently target the most challenging aspects of the feature space, thereby facilitating the comprehensive comprehension and establishing a robust foundation for evaluation.
The second component intends to incrementally capture the evolving patterns within the feature space for enhancing evaluation efficiency.
Specifically, we employ a multi-head attention mechanism as the backbone to develop the evaluator.
The feature subspaces are sequentially fed into the evaluator to capture complex relationships, leveraging contextual information across subspaces.
The evaluator utilizes shared weights across various feature subspaces.
Moreover, since refined feature spaces across consecutive optimization iterations often share overlapping information, we incrementally update the evaluator’s parameters to retain prior evaluation knowledge while incorporating new insights from the evolving feature space.
Finally, we apply \model\ to iterative feature selection (FS) algorithms and conduct extensive experiments on fifteen real-world datasets to validate its superiority and effectiveness.

\section{Related Work}
\textbf{Incremental Learning (IL)} aims to acquire new knowledge without forgetting the knowledge it has already learned \citep{zhu2021class}.
IL is applied in scenarios such as dynamic environments \citep{shieh2020continual,read2012batch} and online learning \citep{shim2021online}. 
IL methods can be divided into three categories: regularization, memory replay, and parameter isolation methods.
Regularization-based methods (\textit{e.g.}, \citep{kirkpatrick2017overcoming,li2017learning}) prevent significant changes in important parameters of previous tasks. 
Memory replay methods retain old task data \citep{isele2018selective} or use generative models to simulate it \citep{shin2017continual}, and then train this data alongside new task data when learning new tasks. 
Parameter isolation methods achieve task isolation by assigning independent model parameters to different tasks (\textit{e.g.}, \citep{rajasegaran2019random,serra2018overcoming}) or by expanding the network structure to accommodate new tasks (\textit{e.g.}, \citep{moriya2018progressive,aljundi2017expert}).
In this paper, we update the evaluator using the Elastic Weight Consolidation (EWC) strategy \citep{kirkpatrick2017overcoming,2021IncDet}. This approach estimates the importance of model parameters for previous tasks and minimizes changes to these important parameters when training on new tasks. This method significantly improves the training efficiency.

\textbf{Multi-Head Attention} mechanism enhances representation capability by simultaneously attending to different subspaces of the input data \citep{vaswani2017attention,messaoud2021trajectory}.
This technique is used in natural language processing \citep{vaswani2017attention,sun2020generating} and object detection \citep{Dai_2021_CVPR} to capture complex patterns and dependencies.  
Unlike previous works, we propose a weighted multi-head attention mechanism that shares weights to encode key characteristics of the feature space into an embedding vector for the evaluation.

\textbf{FS} is widely used in high-dimensional fields \citep{nguyen2020survey,huang2025time}, such as bioinformatics \citep{pudjihartono2022review} and finance \citep{arora2020bolasso}. Among FS methods, the wrapper method stands out for its ability to select features based on model performance \citep{nouri2021novel}. Wrapper-based methods use the performance of the downstream model as a criterion and employ iterative search to find the optimal feature subset \citep{liu2023novel}. 
The representative wrapper method is Recursive Feature Elimination (RFE). RFE iteratively trains the model and removes the least important features, gradually reducing the feature set size until a specified criterion is met \citep{guyon2002gene}.
In this paper, we use FS as a representative example of feature optimization to illustrate the subsequent technical details.

\section{Problem Statement}
This paper introduces a novel feature space evaluator to efficiently identify the optimal feature space. The proposed evaluator can be seamlessly integrated into any iterative feature space optimization algorithm.
Formally, given a dataset $\mathbb{D} = \langle \bm{\mathcal{F}}, \bm{y} \rangle$, where $\bm{\mathcal{F}}$ represents the feature space and $\bm{y}$ denotes the target label space, we first initialize the parameters 
$\bm{\Theta}_{\mathcal{M}}$ of the evaluator $\mathcal{M}$ based on $\mathbb{D}$. This initialization is achieved by minimizing the prediction error $\mathcal{L}$. The learning objective can be defined as:
\begin{equation}
\arg\min_{\bm{\Theta}_{\mathcal{M}}} \mathcal{L}(\mathcal{M}( \bm{\mathcal{F}};{\bm{\Theta}}_{\mathcal{M}}), \bm{y}).
\end{equation} 
In the $t$-th optimization, we can get a new feature space $\langle  \bm{\mathcal{F}}^{(t)}, \bm{y}^{(t)} \rangle$. 
We leverage the information overlap between feature spaces from consecutive iterations to incrementally update $\bm{\Theta}_{\mathcal{M}}^{(t)}$, enabling efficient tracking of evolving patterns and providing an accurate evaluation of the feature space.
The learning process can be formulated as follows:
\begin{equation}
\arg\min_{\bm{\Theta}_{\mathcal{M}}} \mathcal{L}(\mathcal{M}(\mathcal{\bm{\mathcal{F}}}^{(t)};{\bm{\Theta}}_{\mathcal{M}}^{(t)}), \bm{y}^{(t)})+\lambda\Vert \bm{\Theta}_{\mathcal{M}}^{(t)} - \bm{\Theta}_{\mathcal{M}}^{(t-1)} \Vert_2,
\end{equation}
where $\lambda$ is a regularization parameter that balances retaining prior evaluation knowledge with incorporating new insights from the updated feature space, and $\Vert \cdot \Vert_2$ is L2 norm. The learning process continues until either the maximum number of iterations is reached or the optimal feature space is identified.
The design and optimization of $\mathcal{M}$ represent the core contribution of this paper.

\section{Methodology}
\begin{figure*}[t]
    \subcaptionbox{Pipeline framework for \model. \label{fig:framework_a}}{
        \includegraphics[width=12cm,height=5cm]{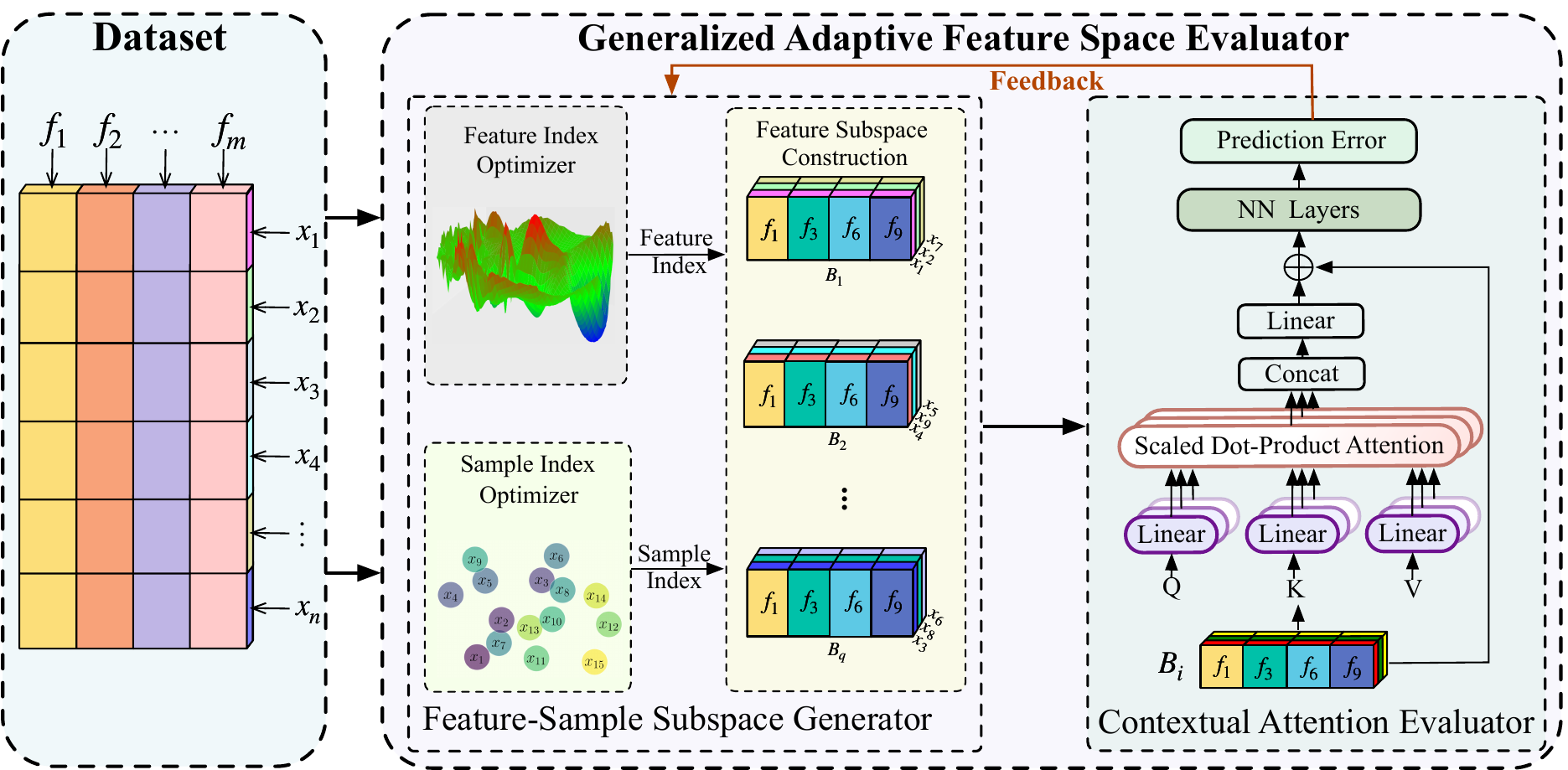}}
    \subcaptionbox{Parameter update. \label{fig:framework_b}}{
        \includegraphics[width=3.5cm,height=5cm]{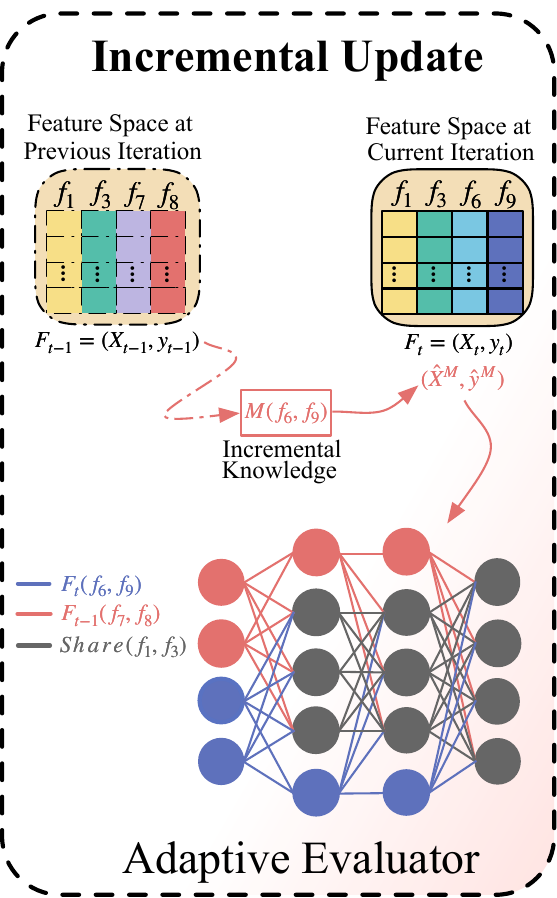}}  
     \caption{Framework overview and parameter update for \model.
    The framework comprises two key components: the Feature-Sample Subspace Generator and the Contextual Attention Evaluator. The first component aims to decouple the complex information within the feature space, enabling the evaluator to focus on capturing the most challenging aspects for evaluation. The second component is designed to comprehensively capture the characteristics of the feature space, ensuring fair and accurate evaluation. By considering information overlap across consecutive iterations, the evaluator incrementally updates its parameters, enhancing the efficiency of the overall optimization process.}
\end{figure*}

\textbf{Framework Overview.}
Figure \ref{fig:framework_a} shows the framework overview of \model, which includes two key components: 1) Feature-Sample Subspace Generator; and 2) Contextual Attention Evaluator.
The first component decouples the information distribution within the evolving feature space, aiming to reduce the complexity for the subsequent evaluator in capturing the key characteristics of the feature space.
Specifically, in each optimization, guided by the evaluator, we first use the feature index optimizer to identify the most important features for the downstream prediction task. Then, the sample index optimizer is used to discover samples that are challenging to evaluate.
After that, we construct fixed-length feature subspaces by applying a random combination strategy to the identified features and samples.
The second component aims to efficiently capture complex feature interactions within the feature space to facilitate effective evaluation.
In detail, we first input the multiple feature subspaces constructed by the first component into the contextual attention evaluator to capture complex interactions and encode them into an embedding vector.
In addition, the embedding vector is used to perform the evaluation task and the resulting prediction error is fed back to the previous component as guidance.
During this process, as illustrated in Figure \ref{fig:framework_b}, considering the partial information overlap between feature spaces in consecutive iterations, we incrementally update the evaluator's parameters to efficiently incorporate new evaluation insights.

\subsection{Feature-Sample Subspace Generator}
\label{sec:feature_subspace}
\textbf{Why is feature space decoupling important?} During the iterative feature space optimization process, complex feature interactions can obscure the underlying patterns. 
A comprehensive understanding of these interactions is crucial for an accurate and objective evaluation. 
By decoupling the feature space, we can reduce the complexity of the learning task, allowing the evaluator to concentrate on the most challenging aspects, ultimately resulting in more effective and precise evaluations. 

\textbf{Feature Index Optimizer} is designed to identify the features most relevant to the subsequent evaluation task. We derive the feature index subset based on the feature importance scores.
Formally, in the $t$-th iteration, given the feature space $\bm{\mathcal{F}}^{(t)}$ and the target $\bm{y}^{(t)}$, the importance score $\text{Score}(\bm{f}_i)$ for each feature $\bm{f}_i$ is calculated by assessing the impact of removing that feature on the performance of the model. The $\text{Score}(\bm{f}_i)$ is computed as follows:
\begin{equation}
    \text{Score}(\bm{f}_i) = \mathcal{M}(\bm{\mathcal{F}}^{(t)}; \bm{\Theta}_{\mathcal{M}}^{(t)}) - \mathcal{M}(\bm{\mathcal{F}}^{(t)} \setminus \{\bm{f}_i\}; \bm{\Theta}_{\mathcal{M}}^{(t)}). \label{feature_index}
\end{equation}
Here, $\mathcal{M}(\cdot; \bm{\Theta}_{\mathcal{M}}^{(t)})$ denotes the feature space evaluator that measures model performance, and $\bm{\mathcal{F}}^{(t)} \setminus \{\bm{f}_i\}$ represents the feature space $\bm{\mathcal{F}}^{(t)}$ with feature $\bm{f}_i$ omitted. Once the importance scores for all features have been computed, they are ranked in descending order. The features with higher ranks are considered to be the most significant contributors to the performance of the model.
We select the best $k$ features based on their importance scores to identify the subset of the feature index
$\bm{f}^{(t)} = \{\bm{f}_1, \bm{f}_2, \cdots, \bm{f}_k\}$ that is the most relevant to the evaluation task.
This component can be replaced with any FS module, allowing \model\ to be compatible with any iterative FS algorithm.

\textbf{Sample Index Optimizer} aims to select the most challenging samples for the subsequent evaluation task.
The sample index subset is derived based on the evaluation error from the feature space evaluator.
Formally, in the $t$-th iteration, the feature space consists of $n$ samples and a target variable $\bm{y}$.
    For the $i$-th sample $x_i$, the prediction error is given by $\mathcal{L}_i^{(t-1)}=\ell(\bm{y}_i,\hat{\bm{y}}_i)$, where $\ell$ denotes the evaluation metric, $\bm{y}_i$ is the target value, and $\hat{\bm{y}}_i$ is the prediction value.
The sampling probability for sample 
$x_i$ is defined as:
$
P(X = x_i) = \frac{\mathcal{L}_i^{(t-1)}}{\sum_{j=1}^{n} \mathcal{L}_j^{(t-1)}}
$,
where $P(X = x_i)$ represents the likelihood of selecting sample $x_i$ based on its relative prediction error. 
This ensures that samples with large errors have a higher probability of being selected.
To efficiently sample from this distribution, we use the cumulative distribution function (CDF), which allows us to transform a uniformly distributed random number into a sample from the desired probability distribution. The CDF is constructed as:
$
C_i = \sum_{k=1}^{i} p_k = \sum_{k=1}^{i} P(X = x_k)
\label{equ:sample_index}
$,
where $C_i $ represents the cumulative sum of probabilities up to the $i$-th sample. 
More specifically, in the weighted sampling process, we first generate a random number $r$ uniformly distributed in the interval $[0,1]$.
Next, we identify the first sample index $i$ such that the CDF satisfies $C_i \geq r$, and select the corresponding sample $x_i$. 
The process is repeated until a new set of sample indices $\mathbb{I}^{(t)}$ is collected.
Using the CDF, the sampling process aligns with the distribution of prediction errors, giving higher priority to samples with larger errors.

\textbf{Feature Subspace Construction} decouples the complex feature space into distinct portions to improve the understanding of the subsequent evaluator for a fair evaluation.
Thus, we introduce the strategy for the construction of feature subspaces using the feature index $\bm{f}^{(t)}$ and sample index $\mathbb{I}^{(t)}$.
Specifically, we sample $s$ sample indices from $\mathbb{I}^{(t)}$ for $q$ times using repeated sampling to obtain various sub sample indices $\{\mathbb{I}^{(t)}_1,\mathbb{I}^{(t)}_2, \cdots, \mathbb{I}^{(t)}_q\}$ with the same length.
Next, we use the obtained sample indices and $\bm{f}^{(t)}$ to select the corresponding samples and features, constructing various feature subspaces denoted as $\mathcal{B}^{(t)} =\{\bm{B}^{(t)}_{1}, \bm{B}^{(t)}_{2}, \cdots, \bm{B}^{(t)}_{q} \}$.
The $i$-th feature subspace $\bm{B}^{(t)}_{i} \in \mathbb{R}^{s\times k}$, where $s$ and $k$ denote the numbers of samples and features in $\bm{B}^{(t)}_{i}$. 
Through this process, we decouple the information distribution within the feature space, preserving the most important and challenging aspects for evaluation in $\mathcal{B}^{(t)}$.

\subsection{Contextual Attention Evaluator}
To thoroughly capture the complicated interactions of the feature space, we design a contextual attention evaluator leveraging a multi-attention mechanism. The learned feature subspaces $\mathcal{B}^{(t)}$, are sequentially fed into the evaluator to facilitate comprehensive information extraction.
We use the $i$-th feature subspace to illustrate the following calculation process. For clarity, we omit the $(t)$ notation for the $i$-th example throughout the remainder of this paper.

Specifically, we begin by projecting $\bm{B}_i$ into three different spaces: the query $\bm{Q}$, key $\bm{K}$, and value $\bm{V}$ spaces.
These projections are computed through linear transformations, which can be defined as:
\begin{equation}
    \bm{Q} = \bm{B}_i\cdot\bm{W}_Q,  \bm{K} = \bm{B}_i\cdot\bm{W}_K,  \bm{V} = \bm{B}_i\cdot\bm{W}_V, 
\end{equation}
where $\bm{W}_Q, \bm{W}_K, \bm{W}_V \in \mathbb{R}^{k \times k}$ are learned weight matrices.
Then, we project them into the subspaces of multi-attention heads. The $h$-th head can be represented through linear projections as:
\begin{equation}
\bm{Q}_h = \bm{Q} \bm{W}^Q_h, \bm{K}_h = \bm{K} \bm{W}^K_h, \bm{V}_h = \bm{V} \bm{W}^V_h,
\end{equation}
where $\bm{W}^Q_h,  \bm{W}^K_h, \bm{W}^V_h\in \mathbb{R}^{k \times d_k}$,  $d_k$ is the dimensionality of  $h$-th head. Then, we compute the attention weights by taking the dot product between the query and key matrices. 
To ensure numerical stability and manageable gradient magnitudes during training, the result is scaled by $\sqrt{d_k}$ and normalized using the softmax function.
These attention weights are used to perform a weighted aggregation of the value matrix $\bm{V}_h$, which can be formulated as: 
\begin{equation}
    \text{Attention}(\bm{Q}_h, \bm{K}_h, \bm{V}_h)   = \text{softmax}\left(\frac{\bm{Q}_h \bm{K}_h^T}{\sqrt{d_k}}\right) \bm{V}_h. 
\end{equation}
To comprehensively capture multiple facets of the feature subspace, we design multiple heads, each with the same structure as described above. These heads generate different attention outputs from different perspectives. The resulting attention outputs are then concatenated and passed through a linear transformation to get $\bm{B}'_i \in \mathbb{R}^{s\times k}$, which can be formulated as
\begin{equation}
   \bm{B}'_i = \text{concat}(\text{head}_1, \text{head}_2, \dots, \text{head}_h) \bm{W}_O,
\end{equation}
where $\bm{W}_O\in\mathbb{R}^{k\times k}$ is the output weight matrix.
After that, we concatenate $\bm{B}'_i$ and $\bm{B}_i$ to form a combined representation, which is then passed through a fully connected layer to generate the prediction $\bm{\hat{y}}_i$. This process can be formulated as follows:
\begin{equation}
    \bm{\hat{y}}_i = \text{FC}(\text{Concat}(\bm{B}'_i; \bm{B}_i)).
\end{equation}
This concatenation allows the evaluator to retain both original and context enhanced feature information for more effective prediction. When different feature subspaces are input into the evaluator, the same structure is used, and the weights are shared across all subspaces. This ensures consistency and promotes generalization by learning common patterns.

\subsection{Optimization}
\textbf{Pre-training.} A well-initialized contextual attention evaluator provides a strong foundation for evaluation, allowing faster convergence and ensuring fair evaluation.
To ensure an effective initialization for the contextual attention evaluator $\mathcal{M}$, we pre-train it using the original feature space $\bm{\mathcal{F}}$ as a foundational basis.
Rather than employing the feature index optimizer and sample index optimizer, we construct the feature subspaces $\mathcal{B}^{(0)}$ by randomly sampling the feature and sample indices.
Then, we subsequently input feature subspace within $\mathcal{B}^{(0)}$ into the evaluator to perform the prediction. 
The optimization objective is to minimize the discrepancy between the predicted and actual values, which can be formulated as:
\begin{equation}
    \mathcal{L}_\text{intial} = \sum_{i=1}^s \mathcal{L}_i (\bm{y}^{(0)}_i,\bm{\hat{y}}^{(0)}_i),
\end{equation}
where $\bm{y}^{(0)}_i$ is the associated target label space of $\bm{B}_i^{(0)}$; $\bm{\hat{y}}^{(0)}_i$ is the predicted target label space; and $s$ is number of feature space within $\bm{B}_i^{(0)}$.
After the model converges, the evaluator is initialized with the parameters $\bm{\Theta}_{\mathcal{M}}^{(0)}$.
The loss function can be tailored to the specific task. For classification, cross-entropy loss is commonly used, whereas for regression, mean squared error is typically employed.

\textbf{Incremental Update.} 
In iterative feature optimization, consecutive iterations often exhibit partial overlap in feature space information.
This motivates us to incrementally update the parameters of the contextual attention evaluator $\mathcal{M}$, enabling faster updates and accelerating the entire feature space optimization process.
Specifically, in the $t$-th iteration, we begin by calculating the Fisher information to assess the importance of parameters based on the previous learning iteration.
Given the feature subspaces $\mathcal{B}^{(t-1)}=\{\bm{B}^{(t-1)}_{1}, \cdots, \bm{B}^{(t-1)}_{q} \}$,
the Fisher information for the $j$-th parameter $\bm{\theta}_j$ in the parameter set $\bm{\Theta}_{\mathcal{M}}^{(t-1)}$ of the contextual attention evaluator is computed as:
\begin{equation}
    \mathcal{G}(\bm{\theta}^{(t-1)}_j) = \frac{1}{s} \sum_{i=1}^s (\nabla_{\bm{\theta}_j} \log p(\bm{y}_i^{(t-1)} \mid \bm{B}_i^{(t-1)} ; \bm{\Theta}_{\mathcal{M}}^{(t-1)})^2,
\end{equation}
where $\mathcal{G}(\bm{\theta}^{(t-1)}_j)$ measures the importance of $\bm{\theta}_j$ based on its contribution to the evaluation task in the $t-1$ iteration.
The term $\nabla_{\bm{\theta}_j} \log p(\bm{y}_i^{(t-1)} \mid \bm{B}_i^{(t-1)} ; \bm{\Theta}_{\mathcal{M}}^{(t-1)})$ represents the logarithmic likelihood gradient with respect to $\bm{\theta}_j$ for conducting evaluation.
A higher value of $\mathcal{G}(\bm{\theta}^{(t-1)}_j)$ indicates that  $\bm{\theta}_j$ is crucial to perform an evaluation of $\mathcal{B}^{(t-1)}$ \citep{grosse2016kronecker}.
After that, we impose constraints on parameter updates during the training of feature subspaces $\mathcal{B}^{(t)}$.
The objective is to prevent forgetting shared evaluation knowledge during parameter updates while incorporating new evaluation insights.
The final loss function in the $t$-th iteration is defined as:
\begin{equation}
    \mathcal{L}_{\text{final}}(\bm{\Theta}^{(t)}_{\mathcal{M}}) = \mathcal{L}_{\mathcal{B}^{(t)}}(\bm{\Theta}^{(t)}_{\mathcal{M}}) + \frac{\lambda}{2} \sum_{j} \mathcal{G}(\bm{\theta}^{(t-1)}_j) \left( \bm{\theta}_j^{(t)} - \bm{\theta}_j^{(t-1)} \right)^2,
\end{equation}
where $\mathcal{L}_{\mathcal{B}^{(t)}}(\bm{\Theta}^{(t)}_{\mathcal{M}})$ represents the prediction loss for the current feature subspaces $\mathcal{B}^{(t)}$; $\bm{\theta}_j^{(t)}$ and $\bm{\theta}_j^{(t-1)}$ are the value of the $j$-th parameter from the parameter set of $\bm{\Theta}^{(t)}_{\mathcal{M}}$ and $\bm{\Theta}^{(t-1)}_{\mathcal{M}}$ respectively;
$\lambda$ is a regularization factor that balances incorporating new evaluation knowledge with preserving shared knowledge.
During the optimization procedure, we minimize $\mathcal{L}_{\text{final}}(\bm{\Theta}^{(t)}_{\mathcal{M}})$ to allow the contextual attention evaluator to capture the dynamics of the feature space, promoting faster convergence and more stable learning.

\section{Experiments}

\begin{table}[t]
\centering
\caption{Summary of the datasets.}
\label{tab:dataset} 
\begin{tabular}{p{2cm} p{0.5cm} p{1cm} p{1cm} p{0.5cm} p{1cm}}
 \toprule
        \small{Dataset}  & \small{R/C} & \small{Samples} & \small{Features} & \small{Classes}   & \small{Source}\\ 
        \midrule
        \small{openml\_607}                  &\small{R}      &\small{1000}       &\small{51}      & --               &\small{OpenML} \\
        \small{openml\_616}                  &\small{R}      &\small{500}         &\small{51}      & --               &\small{OpenML} \\
        \small{openml\_620}                  &\small{R}      &\small{1000}       &\small{26}      & --               &\small{OpenML} \\ 
        \small{openml\_586}                  &\small{R}      &\small{1000}       &\small{26}       & --              &\small{OpenML} \\ 
        \small{airfoil}                      &\small{R}      &\small{1503}       &\small{6}       & --               &\small{OpenML} \\ 
        \small{bike\_share}                 &\small{R}      &\small{10886}     &\small{12}       & --               &\small{OpenML} \\
        \small{wine\_red}                       &\small{C}      &\small{999}         &\small{12}      &\small{6}    &\small{UCI}   \\ 
        \small{svmguide3}                     &\small{C}      &\small{1243}        &\small{22}      &\small{2}    &\small{OpenML} \\
        \small{wine\_white}                   &\small{C}      &\small{4898}        &\small{12}      &\small{7}    &\small{OpenML} \\
        \small{spectf}             &\small{C}      &\small{267}         &\small{45}      &\small{2}    &\small{UCI}   \\ 
        \small{spam\_base}                    &\small{C}      &\small{4601}        &\small{58}      &\small{2}    &\small{OpenML} \\
        \small{mammography}             &\small{C}      &\small{11183}       &\small{7}       &\small{2}     &\small{OpenML} \\
        \small{spam\_base}              &\small{C}      &\small{4601}        &\small{58}      &\small{2}    &\small{OpenML} \\
         \small{AmazonEA}              &\small{C}      &\small{32769}        &\small{9}      &\small{2}    &\small{Kaggle} \\
         \small{Nomao}              &\small{C}      &\small{34465}        &\small{118}      &\small{2}    &\small{UCI} \\
        \bottomrule
\end{tabular}
\end{table}

\subsection{Experimental Setup}

\begin{table*}[t]
    \centering
    \small 
    \renewcommand{\arraystretch}{1}
    \caption{Overall performance comparison. The best results are highlighted in \textbf{bold}, and the second-best results are \underline{underlined}.}
    \label{tab:overall} 
     \scalebox{1}{\begin{tabular}{p{1.6cm} p{0.5cm} p{1.5cm} p{1.8cm}  p{1.8cm} p{1.8cm} p{1.8cm} p{1.8cm}p{1.8cm} } 
        \toprule
        Dataset   &R/C  & Metrics   &\model   &LR  &DT   &GBDT   &RF  &XGB  \\ 
        \midrule
    \multirow{3}{*}{\centering openml\_607}  &\multirow{3}{*}{\centering R}  
     &1-MAE     &\textbf{0.729} $\pm$ 0.028     &0.658 $\pm$ 0.058   &0.710 $\pm$ 0.020   &\underline{0.723} $\pm$ 0.017  &0.699 $\pm$ 0.025   &0.722 $\pm$ 0.023 \\
    &&1-RMSE    &\textbf{0.656} $\pm$ 0.036     &0.561 $\pm$ 0.070    &0.626 $\pm$ 0.019   &0.648 $\pm$ 0.018  &0.602 $\pm$ 0.041   &\underline{0.655} $\pm$ 0.027 \\
     &&$R^2$   &\textbf{0.873} $\pm$ 0.025     &0.805 $\pm$ 0.087    &0.847 $\pm$ 0.033   &\underline{0.863} $\pm$ 0.010  &0.835 $\pm$ 0.026   & 0.863 $\pm$ 0.026  \\                       
     \midrule  
    \multirow{3}{*}{\centering openml\_616} & \multirow{3}{*}{\centering R}
     &1-MAE      &\textbf{0.698} $\pm$ 0.039        &\underline{0.685} $\pm$ 0.035      &0.635 $\pm$ 0.011    &0.679 $\pm$ 0.044   &0.633 $\pm$ 0.037   &0.677 $\pm$ 0.014\\
        &  &1-RMSE       &\textbf{0.611} $\pm$ 0.048        &\underline{0.596} $\pm$ 0.054       & 0.531 $\pm$ 0.023  &0.582 $\pm$ 0.063    & 0.534 $\pm$ 0.038   &0.594 $\pm$ 0.022\\
        &  &$R^2$     &\textbf{0.840} $\pm$ 0.035        &0.833 $\pm$ 0.028       & 0.799 $\pm$ 0.039   & 0.826 $\pm$ 0.044  & 0.791 $\pm$ 0.039   &\underline{0.837} $\pm$ 0.016\\
         \midrule
         
         \multirow{3}{*}{\centering openml\_620} & \multirow{3}{*}{\centering R} 
             &1-MAE   &\textbf{0.703} $\pm$ 0.017     &0.629 $\pm$ 0.083    &0.696 $\pm$ 0.014        &0.698 $\pm$ 0.027   &0.687 $\pm$ 0.015  &\underline{0.702} $\pm$ 0.011 \\  
             
        &  &1-RMSE  &\textbf{0.628} $\pm$ 0.021         & 0.524 $\pm$ 0.100    & 0.524 $\pm$ 0.100      &\underline{0.620} $\pm$ 0.033    & 0.605 $\pm$ 0.016  &0.617 $\pm$ 0.022   \\
        &  &$R^2$    &\textbf{0.861} $\pm$ 0.014   & 0.780 $\pm$ 0.072    &0.852 $\pm$ 0.013    & 0.848 $\pm$ 0.031              & 0.846 $\pm$ 0.007  &\underline{0.855} $\pm$ 0.021 \\   
         \midrule  
         
         \multirow{3}{*}{\centering  openml\_586} & \multirow{3}{*}{\centering R} 
        &1-MAE       &\textbf{0.723} $\pm$ 0.011     & 0.687 $\pm$ 0.057       & 0.706 $\pm$ 0.024            & 0.709 $\pm$ 0.024     &0.717 $\pm$ 0.020   &\underline{0.719} $\pm$ 0.021   \\  
        &  &1-RMSE  &\textbf{0.643} $\pm$ 0.019     & 0.595 $\pm$ 0.074      &0.632 $\pm$ 0.033  & 0.627 $\pm$ 0.033     &0.631 $\pm$ 0.025              &\underline{0.639} $\pm$ 0.027          \\
        &  &$R^2$  &\textbf{0.875} $\pm$ 0.015     & 0.818 $\pm$ 0.062       & 0.862 $\pm$ 0.033            & 0.861 $\pm$ 0.022     &0.862 $\pm$ 0.022  &\underline{0.872} $\pm$ 0.010     \\ 
         \midrule    
                \multirow{4}{*}{\centering  mammography} & \multirow{4}{*}{\centering C} 
         &Accuracy    &\textbf{0.989} $\pm$ 0.003            & 0.978 $\pm$ 0.004       & 0.984 $\pm$ 0.004    &\underline{0.987} $\pm$ 0.002      &0.985 $\pm$ 0.005 &0.985 $\pm$ 0.002  \\     
         &  &Precision     & 0.947 $\pm$ 0.042    & 0.837 $\pm$ 0.036       &\textbf{0.976} $\pm$ 0.021    & 0.949 $\pm$ 0.018          & 0.947 $\pm$ 0.010 &\underline{0.953} $\pm$ 0.024 \\
         &  &F1              &\textbf{0.818} $\pm$ 0.041            & 0.653 $\pm$ 0.035       & 0.713 $\pm$ 0.095    & \underline{0.803} $\pm$ 0.041    & 0.757 $\pm$ 0.034  &0.736 $\pm$ 0.028 \\
         &  &Recall       & \textbf{0.831} $\pm$ 0.043           & 0.603 $\pm$ 0.029       & 0.651 $\pm$ 0.078    & \underline{0.733} $\pm$ 0.045    & 0.685 $\pm$ 0.035  &0.662 $\pm$ 0.024\\  
    \midrule
     \multirow{4}{*}{\centering spectf} & \multirow{4}{*}{\centering C} 
         &Accuracy    &\textbf{0.825} $\pm$ 0.041    &0.737 $\pm$ 0.028      &0.781 $\pm$ 0.049     &0.795 $\pm$ 0.055    &\underline{0.815} $\pm$ 0.057  &0.766 $\pm$ 0.033\\     
     &  &Precision   &\underline{0.631} $\pm$ 0.248  &0.467 $\pm$ 0.209      &\textbf{0.643} $\pm$ 0.237    &0.466  $\pm$ 0.147        &0.459 $\pm$ 0.117 &0.482 $\pm$ 0.197   \\
         &  &F1        &\textbf{0.543} $\pm$ 0.054   &0.454 $\pm$ 0.068      &\underline{0.512} $\pm$ 0.077    & 0.478 $\pm$ 0.079      & 0.473 $\pm$ 0.058  &0.450 $\pm$ 0.033 \\
         &  &Recall     &0.515 $\pm$ 0.092          &0.518 $ \pm$ 0.036     &\textbf{0.542} $\pm$ 0.038          &\underline{0.522} $\pm$ 0.044            & 0.514 $\pm$ 0.028    &0.509 $\pm$ 0.018  \\  
      \midrule   
\multirow{4}{*}{\centering AmazonEA} & \multirow{4}{*}{\centering C} 
         &Accuracy   &\textbf{0.963} $\pm$ 0.003    &0.941 $\pm$ 0.004     &\underline{0.944} $\pm$ 0.002                &0.943 $\pm$ 0.002                &0.944 $\pm$ 0.003       &0.942 $\pm$ 0.003\\     
     &  &Precision   &\underline{0.782} $\pm$ 0.199  &0.670 $\pm$ 0.246     &0.772 $\pm$ 0.245                &0.489 $\pm$ 0.001                &\textbf{0.872} $\pm$ 0.199       &0.611 $\pm$ 0.195\\
         &  &F1     &\textbf{0.500} $\pm$ 0.001     &0.488 $\pm$ 0.005     &0.489 $\pm$ 0.003    &\underline{0.490} $\pm$ 0.002                &0.489 $\pm$ 0.003       &0.488 $\pm$ 0.003\\
         &  &Recall   &\textbf{0.503} $\pm$ 0.001             &0.501 $\pm$ 0.002     &0.502 $\pm$ 0.002       &\underline{0.502} $\pm$ 0.001    &\underline{0.502} $\pm$ 0.001       &0.501 $\pm$ 0.002\\ 
    \midrule 
    \multirow{4}{*}{\centering Nomao} & \multirow{4}{*}{\centering C} 
         &Accuracy   &\textbf{0.952} $\pm$ 0.005         &0.942 $\pm$ 0.002   &\underline{0.944} $\pm$ 0.004   &0.940 $\pm$ 0.003   &0.941 $\pm$ 0.003      &0.937 $\pm$ 0.001 \\     
     &  &Precision   &\textbf{0.947} $\pm$ 0.004          &0.940 $\pm$ 0.004   &\underline{0.941} $\pm$ 0.003   &0.937 $\pm$ 0.002   &0.936 $\pm$ 0.003      &0.935 $\pm$ 0.002\\
         &  &F1      &\textbf{0.936} $\pm$ 0.007          &0.927 $\pm$ 0.003   &\underline{0.930} $\pm$ 0.005   &0.924 $\pm$ 0.004   &0.926 $\pm$ 0.004      &0.922 $\pm$ 0.002\\
         &  &Recall   &\textbf{0.922} $\pm$ 0.006       &0.916 $\pm$ 0.003   &\underline{0.920} $\pm$ 0.006   &0.914 $\pm$ 0.006   &0.917 $\pm$ 0.005      &0.911 $\pm$ 0.002\\      
  \midrule 
     \multirow{4}{*}{\centering  wine\_red} & \multirow{4}{*}{\centering C} 
        &Accuracy          &\textbf{0.637} $\pm$ 0.018  &\underline{0.617} $\pm$ 0.042   &0.599 $\pm$ 0.043  &0.596 $\pm$ 0.026   &0.613 $\pm$ 0.020            &0.588 $\pm$ 0.046       \\     
         &  &Precision  &\textbf{0.386} $\pm$ 0.062    &\underline{0.311} $\pm$ 0.099   &0.296 $\pm$ 0.018  &0.264 $\pm$ 0.036   &0.307 $\pm$ 0.066             &0.285 $\pm$ 0.136 \\
         &  &F1          &\textbf{0.332} $\pm$ 0.057    & 0.260 $\pm$ 0.041              &0.290 $\pm$ 0.017  &0.263 $\pm$ 0.027   &\underline{0.300} $\pm$ 0.045  &0.269 $\pm$ 0.088  \\
         &  &Recall       &\textbf{0.334} $\pm$ 0.060    & 0.276 $\pm$ 0.038             &0.294 $\pm$ 0.015  &0.276 $\pm$ 0.025  & \underline{0.308} $\pm$ 0.031  & 0.283 $\pm$ 0.071 \\  
    \bottomrule   
    \end{tabular}}
\end{table*}

\textbf{Datasets.}

\begin{figure*}[t]
    \subcaptionbox{svmguide3}{
        \includegraphics[width=3.4cm,height=3.2cm]{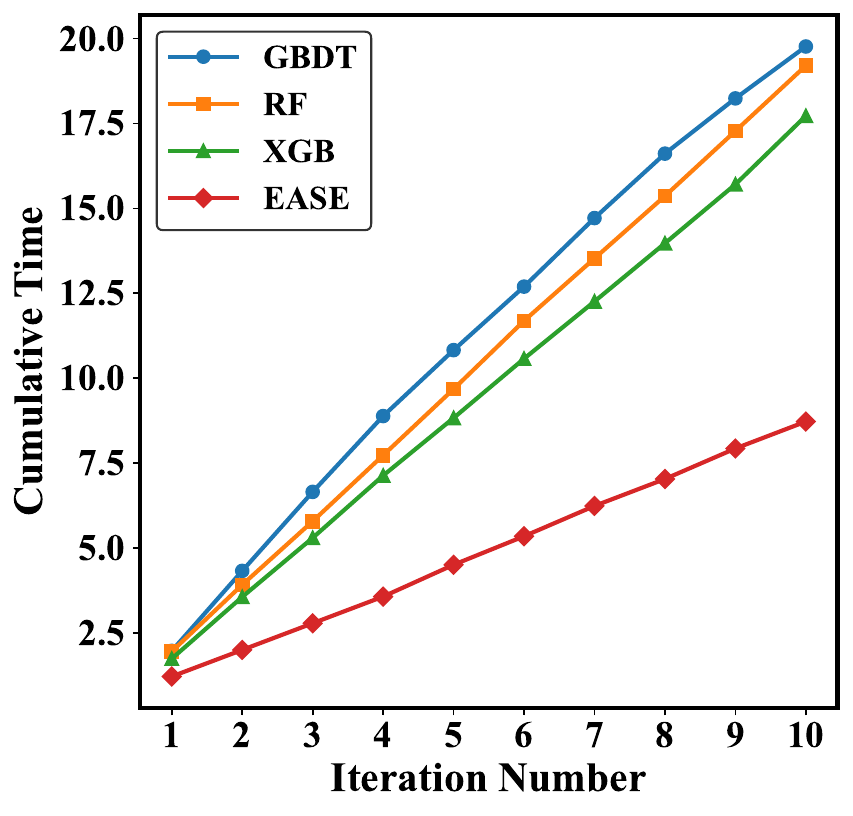}} 
    \subcaptionbox{mammography}{
        \includegraphics[width=3.4cm,height=3.2cm]{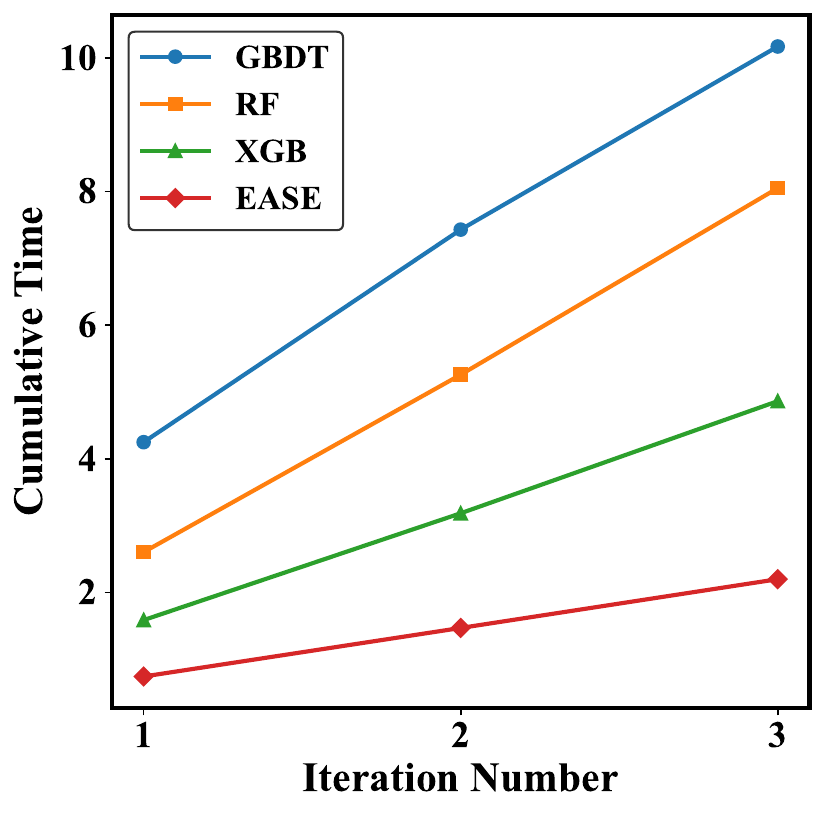}}  
    \subcaptionbox{openml\_620}{
        \includegraphics[width=3.4cm,height=3.2cm]{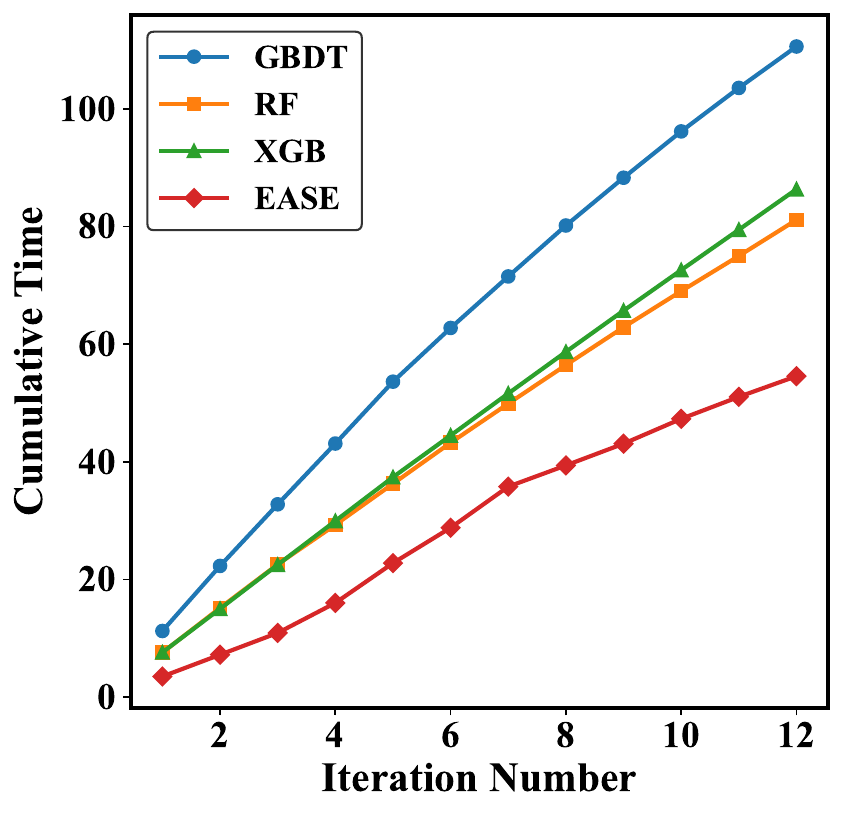}}  
    \subcaptionbox{bike\_share}{
        \includegraphics[width=3.4cm,height=3.2cm]{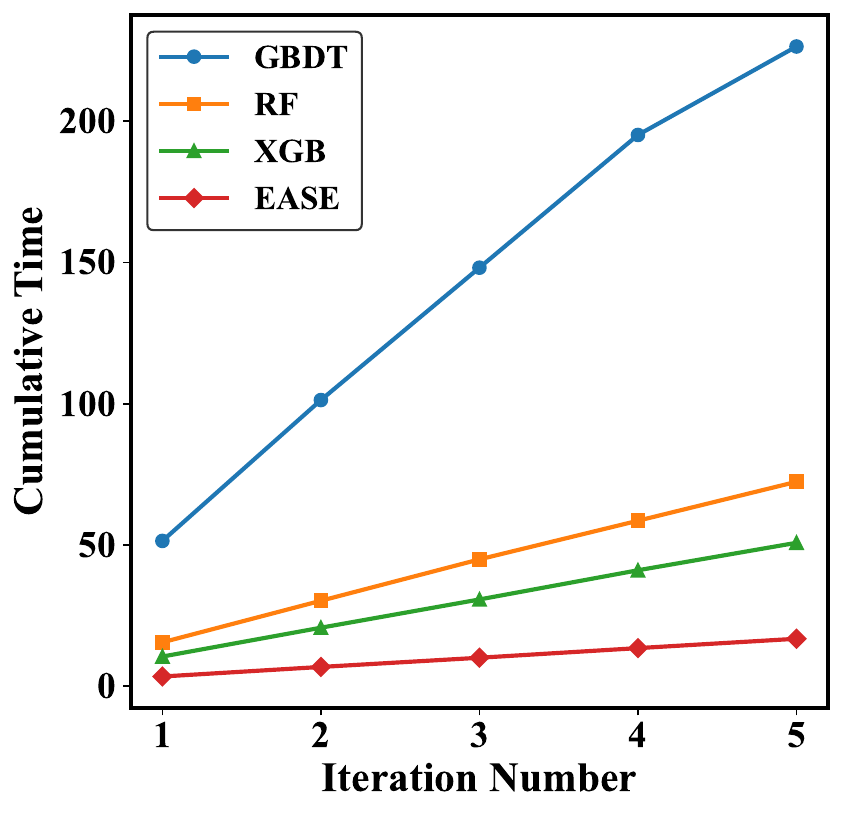}}     
    \subcaptionbox{AmazonEA}{
        \includegraphics[width=3.4cm,height=3.2cm]{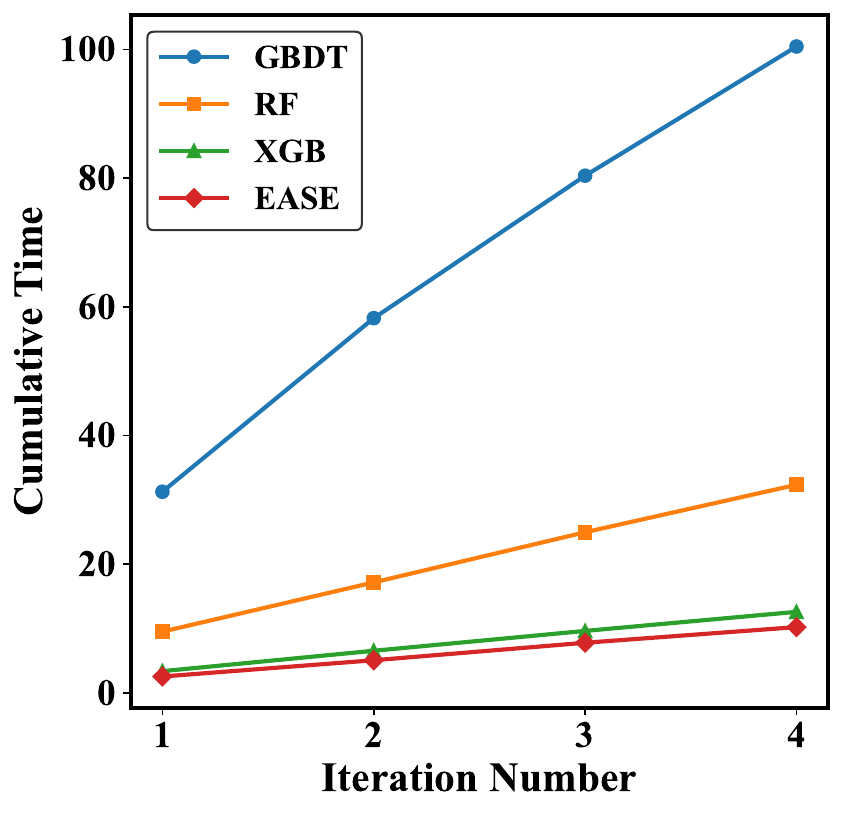}}      
    \subcaptionbox{wine\_white}{
        \includegraphics[width=3.4cm,height=3cm]{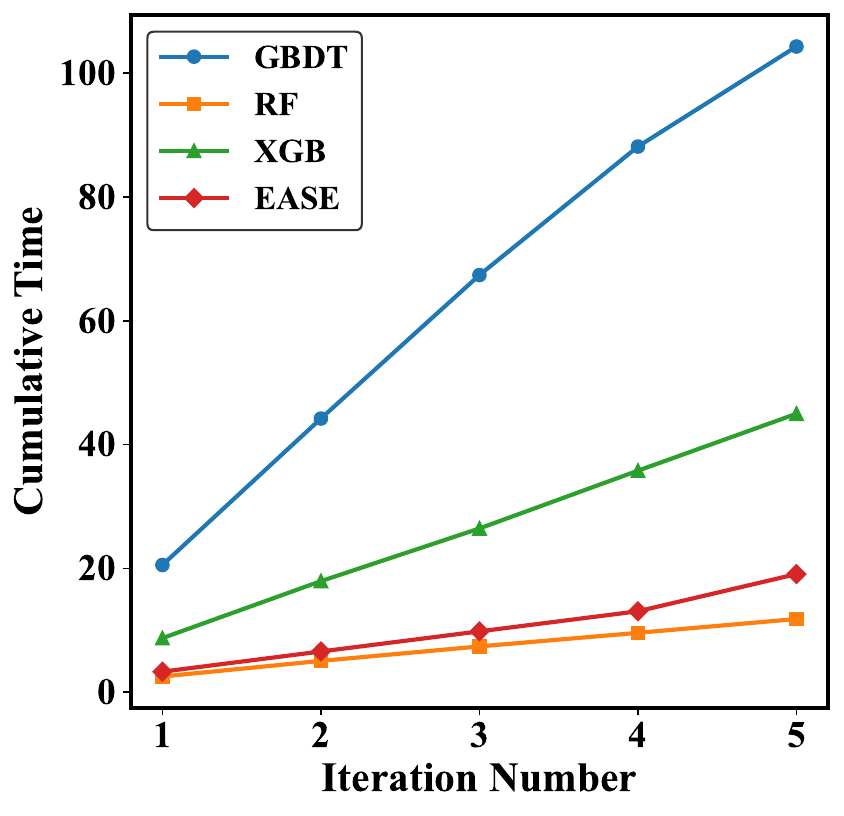}}  
    \subcaptionbox{airfoil}{
        \includegraphics[width=3.4cm,height=3cm]{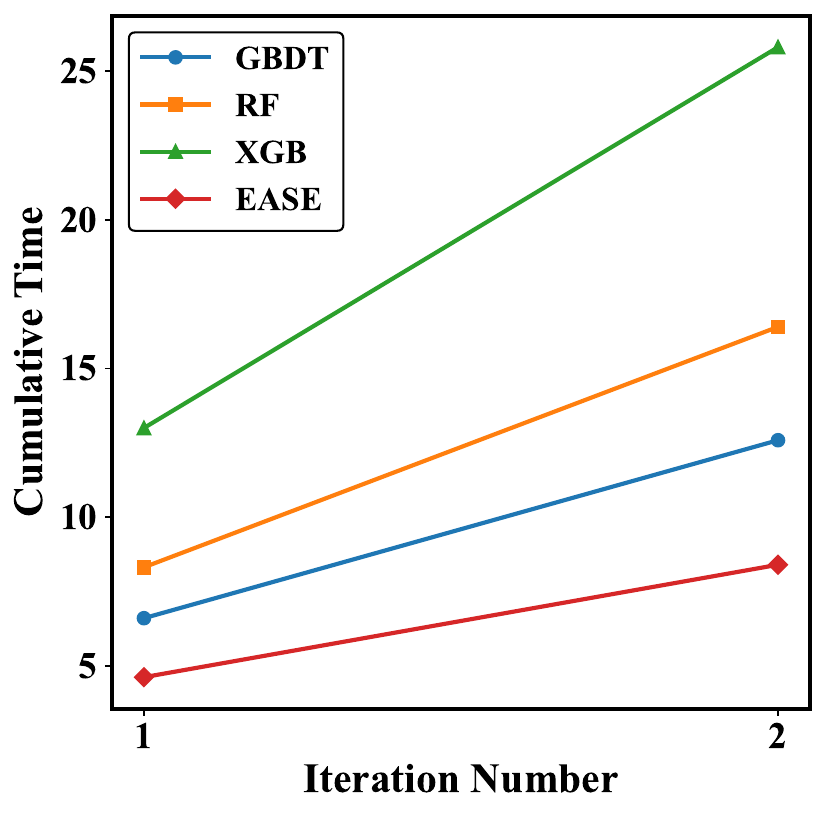}}      
    \subcaptionbox{openml\_616}{
        \includegraphics[width=3.4cm,height=3cm]{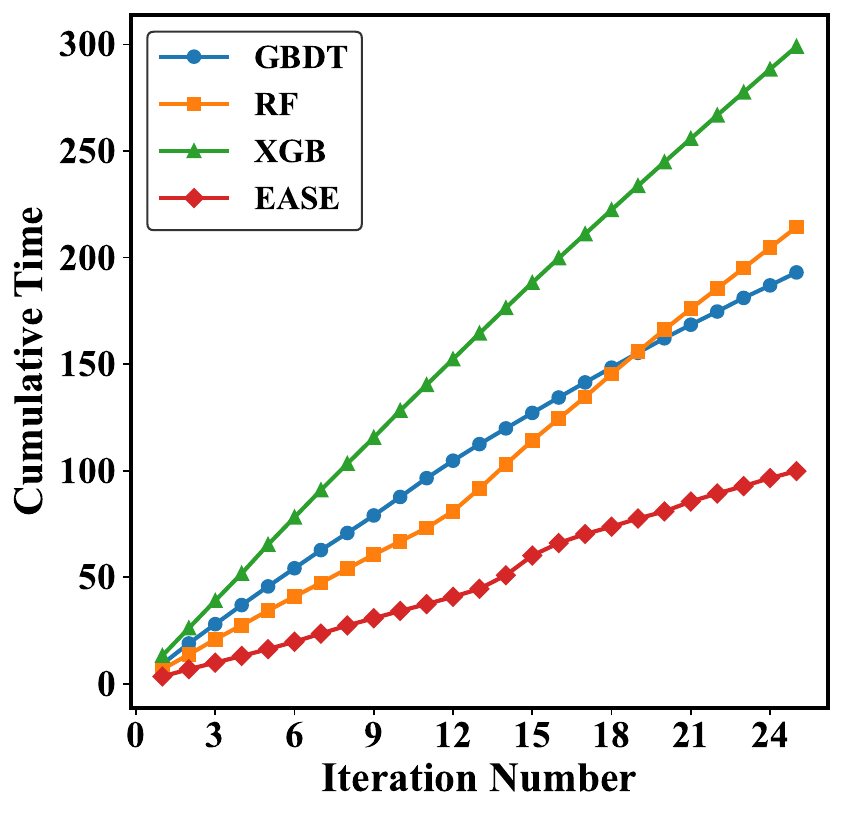}}  
    \subcaptionbox{openml\_607}{
        \includegraphics[width=3.4cm,height=3cm]{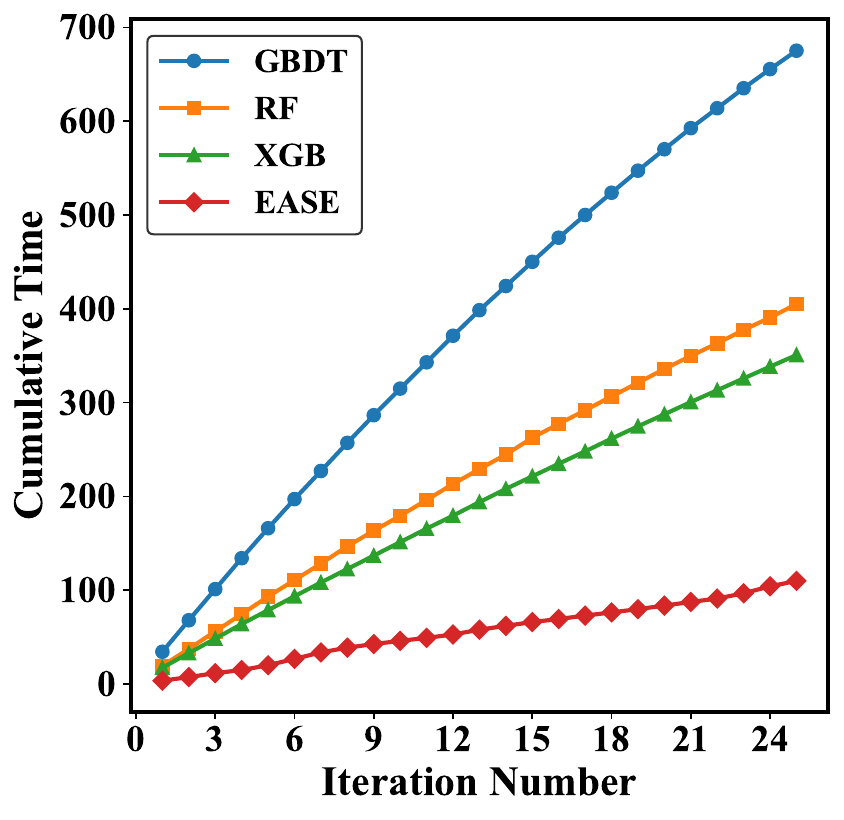}}  
    \subcaptionbox{openml\_586}{
        \includegraphics[width=3.4cm,height=3cm]{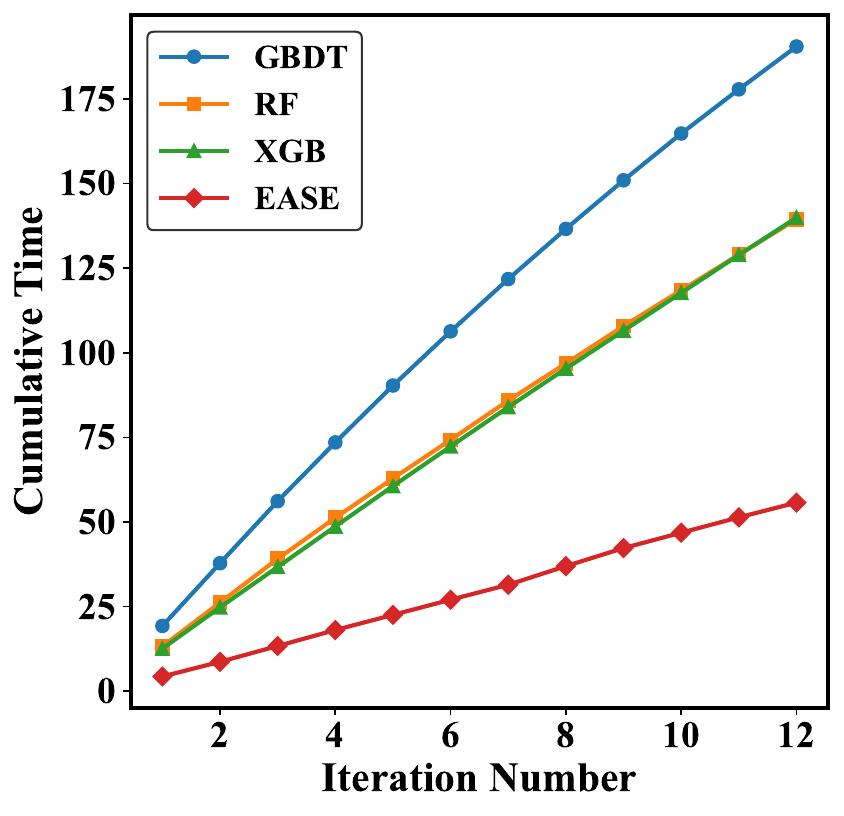}}  
    \caption{Time complexity comparison of different feature space evaluators across various datasets.} 
    \label{fig:time}
\end{figure*}

\begin{figure*}[t]
    \subcaptionbox{spam\_base}{
        \includegraphics[width=3.4cm,height=3.2cm]{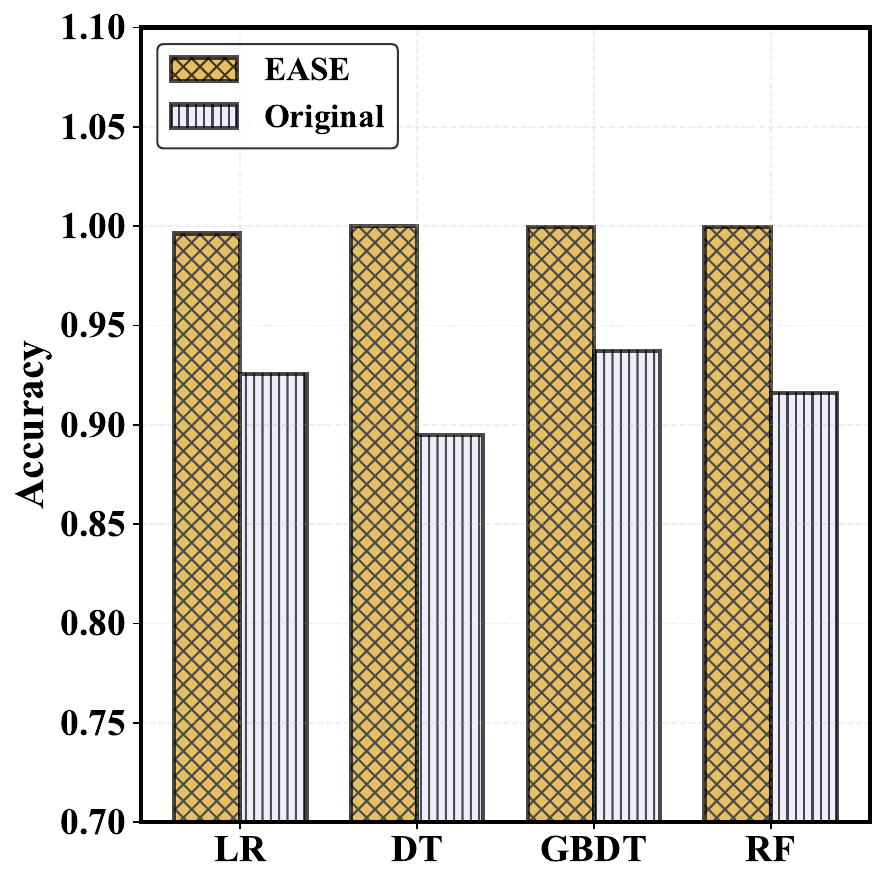}} 
    \subcaptionbox{spectf}{
        \includegraphics[width=3.4cm,height=3.2cm]{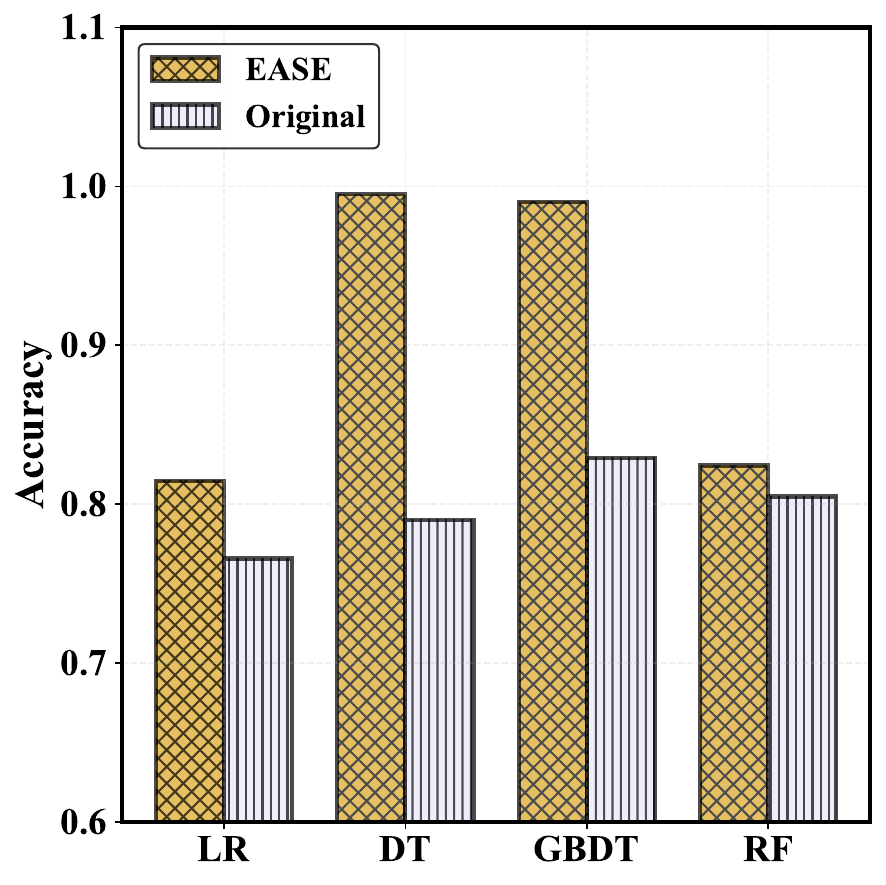}}  
    \subcaptionbox{svmguide3}{
        \includegraphics[width=3.4cm,height=3.2cm]{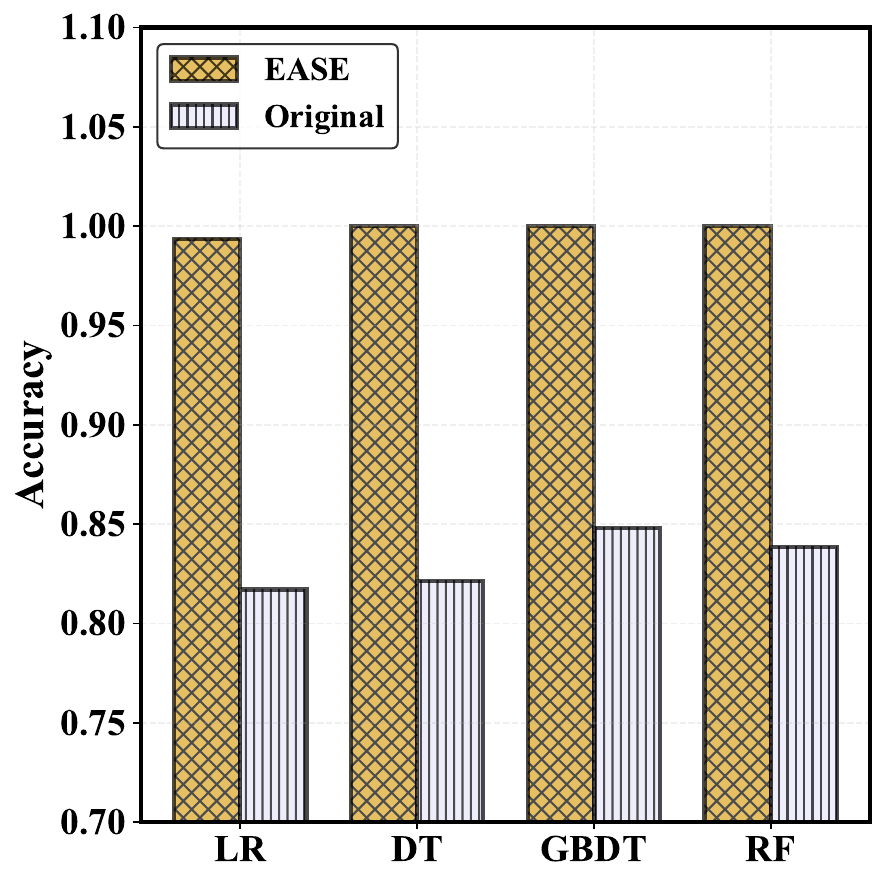}}  
        \subcaptionbox{openml\_616}{
        \includegraphics[width=3.4cm,height=3.2cm]{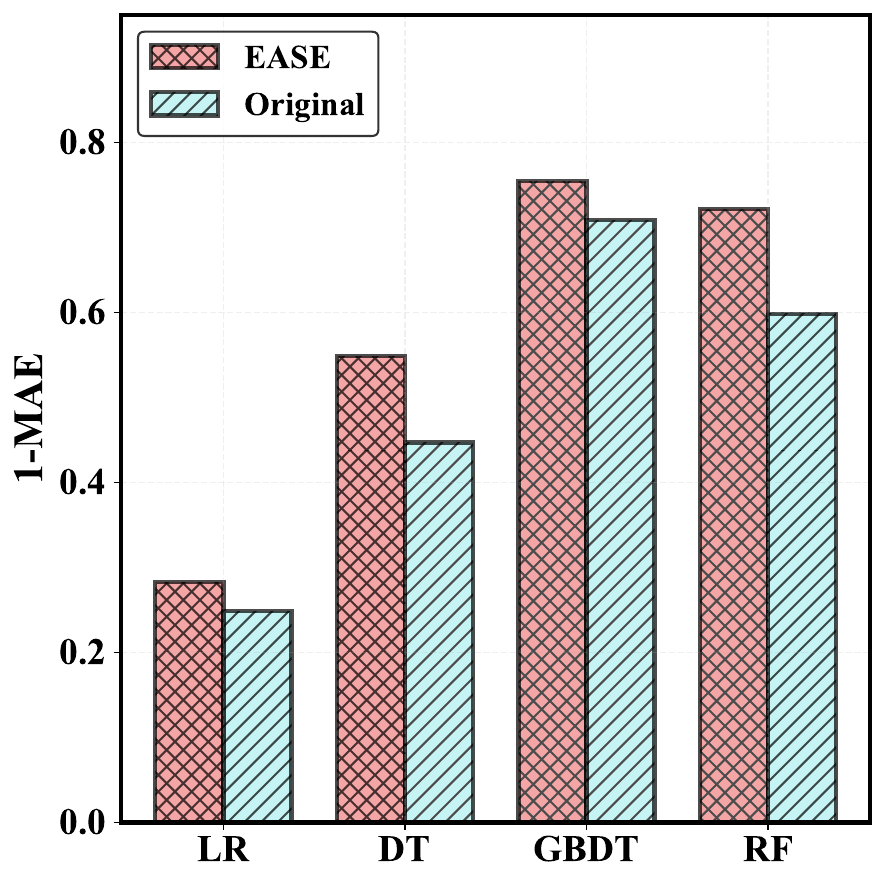}}    
    \subcaptionbox{openml\_607}{
        \includegraphics[width=3.4cm,height=3.2cm]{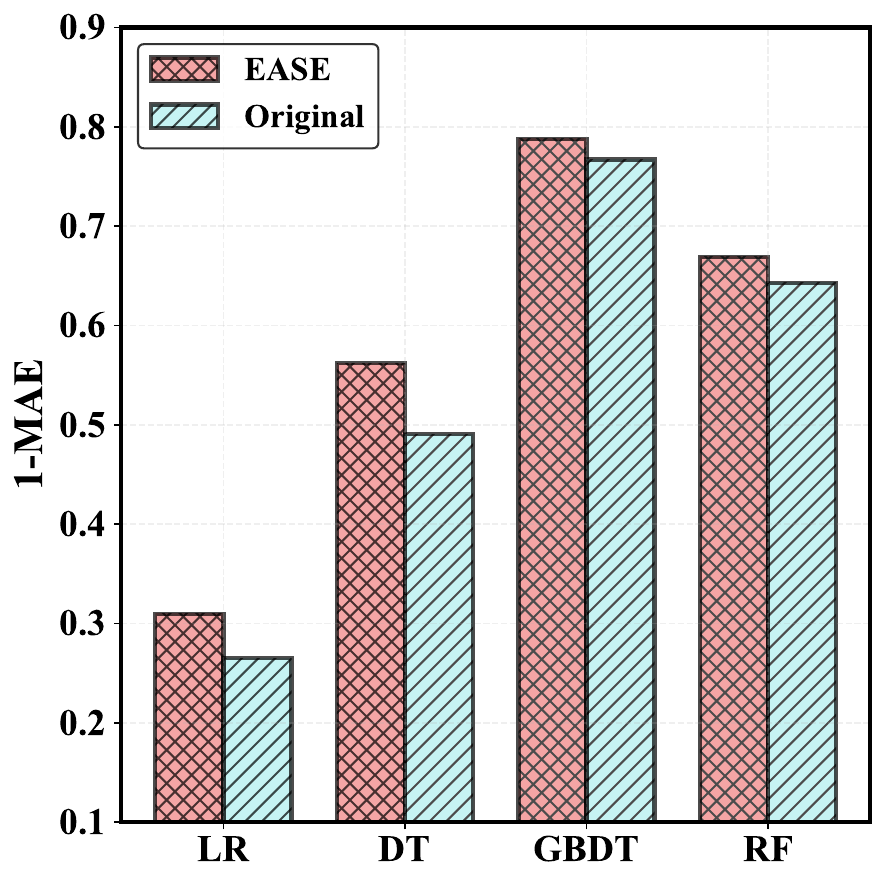}}    
    \caption{Comparison of prediction performance between original and \model\ refined feature spaces.} 
    \label{fig:EASE_Original} 
\end{figure*}

We conduct extensive experiments on 15 publicly available datasets from UCI \citep{openml2024}, OpenML \citep{uci2024} and Kaggle \citep{kaggle2024}, consisting of 9 classification tasks and 6 regression task. A statistical overview of these datasets is presented in Table~\ref{tab:dataset}.
In this table, 'C' denotes classification task datasets, 'R' indicates regression task datasets, and '-' indicates not applicable.

\textbf{Evaluation Metrics.}
We use Mean Absolute Error (MAE), Root Mean Squared Error (RMSE), and R-squared ($R^2$) to evaluate the performance of regression tasks, and Accuracy, Precision, Recall, and F1 score to evaluate the performance of classification tasks.

\textbf{Baseline Algorithms.}
We apply \model\ to three iterative feature optimization frameworks, including both the most classical and state-of-the-art approaches, to validate its effectiveness and generalization capability:
(\romannumeral 1) \textbf{RFE} \citep{guyon2002gene} iteratively eliminates the least important features from the original set until a stopping criterion is met. 
Unless otherwise noted, we employ RFE as the default feature optimization framework throughout our experiments, serving as the baseline for comparing the performance of different evaluators.
 (\romannumeral 2) \textbf{SDAE} \citep{hassanieh2024selective} is a state-of-the-art algorithm designed to select features used in unlabeled datasets without compromising information quality. 
(\romannumeral 3) \textbf{GRFG} is a Feature Generation (FG) method \citep{wang2022group} that addresses challenges in representation space reconstruction via a cascading deep RL approach, employing a group-wise strategy and nested interactive processes to automate FG.

Additionally, we employ six widely-used ML algorithms as feature space evaluators to compare their experimental performance against \model: (1) Linear Regression / Logistic Regression (\textbf{LR}): Linear Regression \citep{su2012linear} models a linear relationship between the features and labels. Logistic Regression \citep{nusinovici2020logistic} classifies data by linearly combining input features and applying a logistic function to the result. 
\textbf{LR} refers to linear regression for regression tasks and logistic regression for classification tasks.
(2) Decision Tree (\textbf{DT}): \citep{kim2014predicting} is a tree-like structure method, used to predict data through some rules. 
(3) Gradient Boosting Decision Tree (\textbf{GBDT}): \citep{li2023interpolation} builds an ensemble of decision trees sequentially to minimize errors. 
(4) Random Forest (\textbf{RF}): \citep{khajavi2023predicting} is an ensemble learning method that constructs multiple decision trees. (5) Extreme Gradient Boosting (\textbf{XGB}) \citep{asselman2023enhancing} combines the strengths of gradient boost with regularization techniques.
In the testing phase, we use RF in all cases to report the performance of the refined feature space.

\textbf{Implementation Details.}
All experiments were conducted on the macOS Sonoma 14.0 operating system, Apple M3 Chip with 8 cores (4 performance and 4 efficiency), and 8GB of RAM, with the framework of Python 3.8.19 and TensorFlow 2.13.0.
We limit pre-training and incremental training to 50 and 200, respectively. We employ an early stopping strategy, stopping the training  when the loss does not decrease for 10 consecutive epochs.
we use the Adam optimizer and set the learning rate starting from $0.001$ and decay by $0.1$ rate every 30 epochs, a batch size of 4, and 16 attention heads.
In all experiments, we use the Adam optimizer and a learning rate decay strategy to accelerate the convergence. All experiments run 10 times and calculate the value of  mean and standard deviation.

\begin{table*}[h]
\centering  
\caption{Comparison of different evaluators in terms of Accuracy (for classification tasks) and 1-MAE (for regression tasks) in SDAE framework. The best results are highlighted in \textbf{bold}. The second-best results are highlighted in \underline{underline}. } 
\label{tab:sdae}  
\scalebox{1}{\begin{tabular}{l l c c c c c c}  
\toprule  
Dataset       & R/C      &\model       & LR          &DT        &GBDT       &RF       &XGB  \\
\midrule 
openml\_607   & R    &\textbf{0.768} $\pm$ 0.034   &0.260 $\pm$ 0.040   &0.603 $\pm$ 0.030   &\underline{0.767} $\pm$ 0.015   &0.722 $\pm$ 0.012   &0.757 $\pm$ 0.011   \\

openml\_616   & R    &\textbf{0.798} $\pm$ 0.014  &0.236 $\pm$ 0.045  &0.482 $\pm$ 0.024   &\underline{0.686} $\pm$ 0.026   &0.657 $\pm$ 0.030   &0.669 $\pm$ 0.033      \\

openml\_620   & R    &\textbf{0.801} $\pm$ 0.005  &0.265 $\pm$ 0.033  &0.523 $\pm$ 0.007   &0.673 $\pm$ 0.013   &0.642 $\pm$ 0.024   &\underline{0.676} $\pm$ 0.012     \\

openml\_586   & R    &\underline{0.772} $\pm$ 0.016  &0.287 $\pm$ 0.041  &0.617 $\pm$ 0.008   &\textbf{0.773} $\pm$ 0.019   &0.736 $\pm$ 0.019   &0.771 $\pm$ 0.010      \\

airfoil       & R    &\textbf{0.846} $\pm$ 0.013  &0.436 $\pm$ 0.025  &0.727 $\pm$ 0.012   &0.714 $\pm$ 0.017   &0.805 $\pm$ 0.008   &\underline{0.838} $\pm$ 0.012     \\

bike\_share   & R    &\textbf{0.993} $\pm$ 0.000  &0.980 $\pm$ 0.000  &0.984 $\pm$ 0.001   &0.981 $\pm$ 0.000   &\underline{0.993} $\pm$ 0.001   &0.987 $\pm$ 0.001     \\

wine\_red     & C    &\textbf{0.741} $\pm$ 0.037      &0.591 $\pm$ 0.008    &0.549 $\pm$ 0.022   &0.640 $\pm$ 0.020     &\underline{0.724} $\pm$ 0.032     &0.600 $\pm$ 0.009   \\        
svmguide3    & C    &\underline{0.856} $\pm$ 0.014   &0.806 $\pm$ 0.021    &0.795 $\pm$ 0.018   &0.832 $\pm$ 0.024     &\textbf{0.861} $\pm$ 0.012        &0.851 $\pm$ 0.004   \\
wine\_white   & C    &\textbf{0.686} $\pm$ 0.034      &0.518 $\pm$ 0.011    &0.585 $\pm$ 0.010   &0.584 $\pm$ 0.009     &\underline{0.675} $\pm$ 0.015     &0.654 $\pm$ 0.008   \\
spectf       & C    &\underline{0.829} $\pm$ 0.046   &0.789 $\pm$ 0.064    &0.772 $\pm$ 0.030   &0.756 $\pm$ 0.072     &\textbf{0.837} $\pm$ 0.057        &0.813 $\pm$ 0.023   \\
mammography  & C    &\textbf{0.992} $\pm$ 0.001      &0.983 $\pm$ 0.001    &0.980 $\pm$ 0.003   &0.986 $\pm$ 0.002     &0.987 $\pm$ 0.004                 &\underline{0.987} $\pm$ 0.001   \\
spam\_base   & C    &\textbf{0.963} $\pm$ 0.006      &0.927 $\pm$ 0.007    &0.904 $\pm$ 0.013   &0.946 $\pm$ 0.003     &\underline{0.960} $\pm$ 0.007     &0.946 $\pm$ 0.005  \\
AmazonEA      & C    &\textbf{0.950} $\pm$ 0.003      &0.944 $\pm$ 0.002    &0.930 $\pm$ 0.003   &0.942 $\pm$ 0.006     &\underline{0.947} $\pm$ 0.002     &0.943 $\pm$ 0.001  \\
Nomao        & C    &\textbf{0.973} $\pm$ 0.002      &0.941 $\pm$ 0.004    &0.945 $\pm$ 0.003   &0.953 $\pm$ 0.002     &0.967 $\pm$ 0.001                  &\underline{0.969} $\pm$ 0.002  \\
\bottomrule 
\end{tabular} }  
\end{table*}

\subsection{Experimental Results}
\subsubsection{Overall Comparison}
\label{sec:overall_comparison}
This experiment aims to answer: \textit{Can \model\ accurately assess feature space quality to produce an effective feature space?} 
We choose RFE as iterative FS framework and adopt \model\ as the feature space evaluator.
To compare the performance difference, we then replace the evaluator with LR, DT, GBDT, RF and XGB, respectively. 
We report the testing performance of the refined feature space using RF.
Table~\ref{tab:overall} shows the comparison results in terms of different  metrics according to different task types.
We can find that \model\ outperforms other baseline algorithms in most cases.
For classification task, \model\ can improve by approximately 3\% compared to other baselines.
For regression task, \model\ demonstrates the most superior performance. 
The underlying driver is that the proposed information decoupling strategy and context-aware evaluator, which allow the evaluator to focus on the most challenging aspects of the feature space.
This results in a fairer evaluation, leading to a more effective refinement strategy and ultimately producing a more optimized feature space.
In summary, this experiment shows that \model\ effectively evaluates feature space quality for better feature spaces.

\subsubsection{Efficiency Comparison}
\label{sec:time_complexity}
This experiment aims to answer: \textit{Is \model\ more efficient compared to other feature space evaluators?}
We compare the training time of \model\ with other feature space evaluators, including GBDT, RF, and XGB.
Figure~\ref{fig:time} shows the comparison results in terms of cumulative time.
An interesting observation is that using \model\ can significantly reduce the cumulative time costs compared to other baselines.
A potential reason is that our incremental parameter update strategy enables the feature space evaluator to quickly capture evolving patterns in the feature space, thereby accelerating the feature optimization process.
To sum up, this experiment demonstrates that \model\ can efficiently assess feature space quality, thanks to its adaptive parameter update strategy.

\subsubsection{The effectiveness of \model\ for Feature Space Refinement}
\label{sec:original_EASE}
This experiment aims to answer: \textit{Is the quality of the feature space refined by \model\ superior to the original feature space?}
We compare the prediction performance between the original feature space and the space refined by \model\ using various downstream predictors, including LR, DT, GBDT, and RF.
Figure~\ref{fig:EASE_Original} shows the comparison results in terms of Accuracy and MAE according to the task type.
We find that the feature space produced by \model\ outperforms the original in most cases across various predictors.
In particular, the refinement by \model\ outperforms the original feature space by 20\% on the spectf dataset.
This observation suggests that incorporating \model\ into the iterative feature space optimization framework provides effective guidance for obtaining a better feature space. 
Additionally, the contextual attention evaluator comprehensively captures the intrinsic traits of the feature space, leading to robust performance across various datasets and predictors.

\subsubsection{\model's Performance in Different FS Iterative Frameworks}
This experiment aims to answer: \textit{Is \model\ generalizable and applicable across different iterative feature space optimization algorithms?}
We apply \model\ to a state-of-the-art FS algorithm SDAE \citep{hassanieh2024selective}. SDAE learns low-dimensional representations of high-dimensional data through a deep auto-encoder structure, while introducing a selective layer that automatically selects a relevant subset of features representing the entire feature space. 
We select \model, LR, DT, GBDT, RF, and XGB as evaluators to evaluate the the selected feature space, respectively . 
Table~\ref{tab:sdae} shows the results across different datasets. We observe that in the feature space selected by the SDAE, \model\ exhibits the best performance in all tasks.

These observations highlight the strong generalizability and applicability of \model.
The underlying driver is that the feature index optimizer of \model\ offers flexibility, allowing it to adapt to various iterative feature optimization frameworks.
In summary, this experiment demonstrates that \model\ exhibits strong adaptability to iterative FS frameworks and excellent generalizability across different feature space optimization algorithms.

\subsubsection{\model's Performance in FG Frameworks.}
\label{app:different_feature_generation}
This experiment aims to answer: \textit{Is \model\ generalizable and applicable in FG algorithm?}
We respectively select \model\, LR, DT, GBDT, RF, and XGB as evaluators to evaluate the performance of the generated features space during the GRFG procedure.
Table~\ref{tab:GRFG} shows the comparison results of six evaluators across different datasets. We observe that the proposed \model\ achieves the best performance in both classification and regression tasks. These results further demonstrates that, compared to traditional feature evaluation algorithms, \model\ exhibits excellent performance in both FS and FG tasks and can capture the key information of the feature space.
\begin{table*}[t]
\centering  
\caption{Comparison of different evaluators in terms of Accuracy (for classification tasks) and 1-MAE
(for regression tasks) in GRFG framework. The best results are highlighted in \textbf{bold}. The second-best results are highlighted in \underline{underline}. } 
\label{tab:GRFG}  
\scalebox{1}{\begin{tabular}{l l c c c c c c}  
\toprule 
Dataset       & R/C        & \model                         & LR                   & DT               &GBDT                   &RF        &XGB  \\
\midrule 
openml\_616   & R      &\textbf{0.698} $\pm$ 0.006      &0.679 $\pm$ 0.002     &0.673 $\pm$ 0.006     &0.669 $\pm$ 0.012      &\underline{0.688} $\pm$ 0.012    & 0.662 $\pm$ 0.000  \\
openml\_586  & R      &\textbf{0.627} $\pm$ 0.001      &0.590 $\pm$ 0.008      &0.590 $\pm$ 0.009    &0.591 $\pm$ 0.003       &\underline{0.615} $\pm$ 0.017    & 0.601 $\pm$ 0.002  \\

svmguide3     & C      &\textbf{0.849} $\pm$ 0.002      &0.816 $\pm$ 0.000      &0.821 $\pm$ 0.006    &0.822 $\pm$ 0.006       &\underline{0.826} $\pm$ 0.000    & 0.820 $\pm$ 0.005  \\
mammography   & C      &\textbf{0.993} $\pm$ 0.001      &0.984 $\pm$ 0.000      &0.985 $\pm$ 0.018    &0.986 $\pm$ 0.001       &0.985 $\pm$ 0.000    &\underline{0.986} $\pm$ 0.000  \\
\bottomrule 
\end{tabular} }  
\end{table*}

\subsubsection{The Impact of Technical Components on Performance}
\label{sec:component}
This experiment aims to answer: \textit{How does each technical component in \model\ impact its performance?}  We investigate the effects of pre-training, incremental training, and feature-sample subspace construction in \model. 
We develop \model$^{-PT}$, \model$^{-IT}$, and \model$^{-FC}$ by removing the pre-training, incremental training, and feature-sample subspace construction components from \model, respectively.
Table~\ref{tab:ablation_performance} shows the results of three different \model\ variants across different datasets. 
We first observe that \model\ outperforms \model$^{-PT}$, highlighting the importance of the pre-training step in providing a strong foundation for feature space evaluation. 
Then, we find that \model\ surpasses \model$^{-IT}$, indicating that the incrementally updating mechanism effectively captures the evolving patterns of feature space optimization, leading to an improved feature space. 
Moreover, experimental results show that \model\ outperforms \model$^{-FC}$, showing that the information decoupling strategy reduces comprehension complexity, allowing for better capture of feature space characteristics and leading to improved evaluation and feature space quality.
In conclusion, this experiment reflects that each technical component in \model\ is indispensable and significant.
\begin{table}[t]
\centering  
\caption{Comparison of different \model\ variants in terms of Accuracy (for classification tasks) and 1-MAE (for regression tasks). The best results are highlighted in \textbf{bold}, and the second-best results are \underline{underlined}. }  
\label{tab:ablation_performance}  
\scalebox{0.7}{\begin{tabular}{p{2cm} p{0.5cm} p{1.8cm}  p{1.8cm} p{1.8cm} p{1.8cm}}   
\toprule 
Dataset      & R/C      & \model                       & \model$^{-FC}$                     & \model$^{-IT}$           &\model$^{-PT}$ \\
\midrule 
openml\_607  &R   &\textbf{0.729} $\pm$ 0.028   &\underline{0.711} $\pm$ 0.028   &0.657 $\pm$ 0.021   &0.687 $\pm$ 0.035  \\

openml\_616  &R    &\textbf{0.698} $\pm$ 0.039   &\underline{0.661} $\pm$ 0.035   &0.565 $\pm$ 0.038   &0.439 $\pm$ 0.082  \\

openml\_620  &R    &\textbf{0.703} $\pm$ 0.017   &0.480 $\pm$ 0.018             &0.617 $\pm$ 0.018  &\underline{0.682} $\pm$ 0.017  \\

openml\_586  &R    &\textbf{0.723} $\pm$ 0.011   &0.652 $\pm$ 0.016          &\underline{0.721} $\pm$ 0.014   &0.636 $\pm$ 0.005\\

airfoil      &R   &\textbf{0.663} $\pm$ 0.009    &0.249 $\pm$ 0.029         &0.288 $\pm$ 0.028        &\underline{0.638} $\pm$ 0.004 \\

bike\_share  &R   &\textbf{0.967} $\pm$ 0.001   &0.965 $\pm$ 0.002    &\underline{0.966} $\pm$ 0.001      &\underline{0.966} $\pm$ 0.001\\

wine\_red      &C     &\textbf{0.637} $\pm$ 0.018   &\underline{0.589} $\pm$ 0.055   &0.539 $\pm$ 0.043         &0.539 $\pm$ 0.028  \\
svmguide3      &C     &\textbf{0.846} $\pm$ 0.034       &0.817 $\pm$ 0.032                  &0.828 $\pm$ 0.014                     &\underline{0.833} $\pm$ 0.010   \\
wine\_white  &C     &\textbf{0.578} $\pm$ 0.017        &0.557 $\pm$ 0.019                & \underline{0.576} $\pm$ 0.014                    &0.552 $\pm$ 0.011\\
spam\_base   &C     &\textbf{0.939} $\pm$ 0.008      &0.918 $\pm$ 0.016                  &0.928 $\pm$ 0.007                  &\underline{0.931} $\pm$ 0.003   \\
mammography  &C     &\textbf{0.989} $\pm$ 0.003       &0.981 $\pm$ 0.004                  &0.985 $\pm$ 0.004                &\underline{0.986} $\pm$ 0.003  \\
spectf       &C    &\textbf{0.825}  $\pm$ 0.041       &\underline{0.815} $\pm$  0.065                 &0.756 $\pm$  0.064                    &0.781 $\pm$  0.041 \\
\bottomrule 
\end{tabular}}  
\end{table} 

\subsubsection{Parameter Sensitivity Analysis}
\begin{figure}[t]
    \subcaptionbox{$D$ on openml\_586 (R).}{
        \includegraphics[width=3.8cm,height=3.6cm]{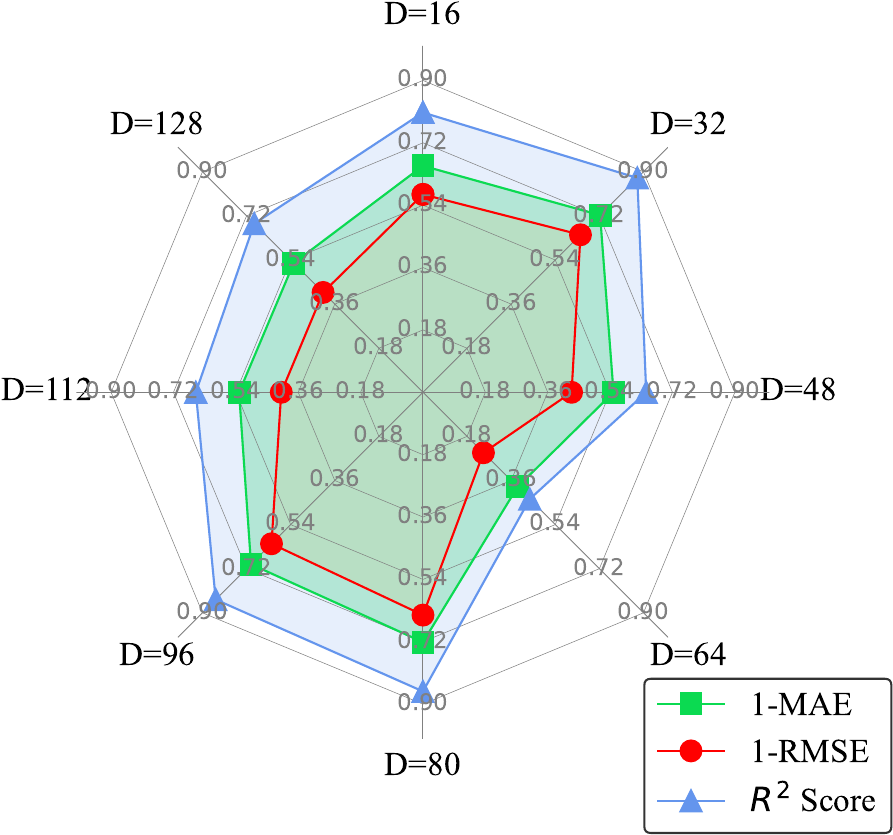}} 
    \subcaptionbox{$h$ on openml\_586 (R).}{
        \includegraphics[width=3.8cm,height=3.6cm]{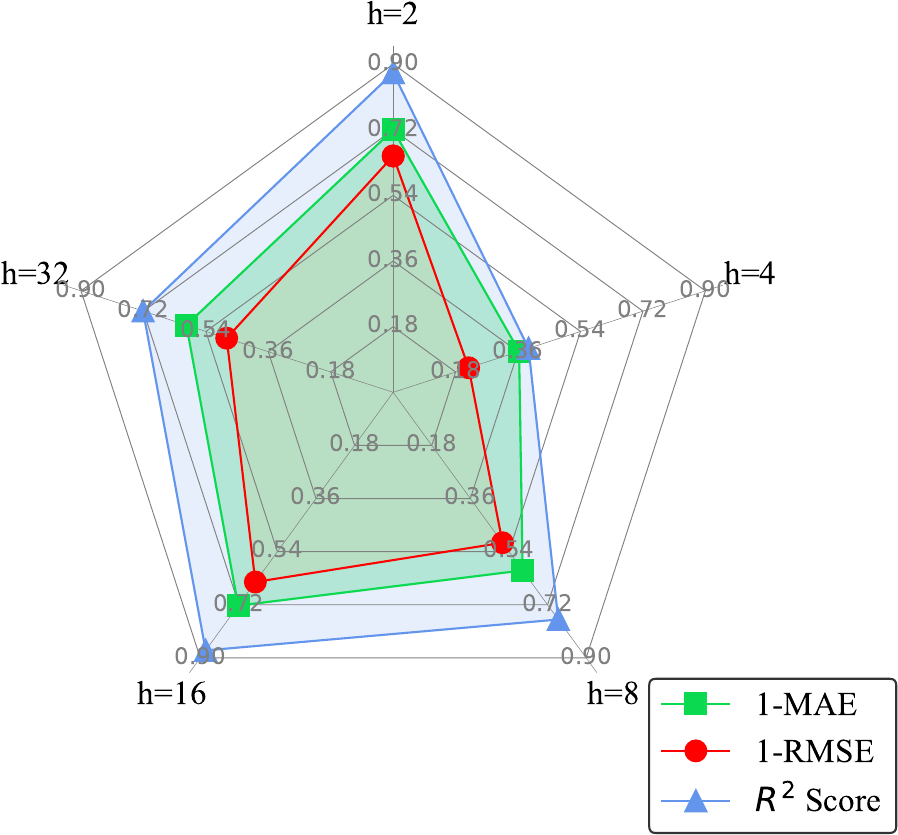}}\vfill 
    \subcaptionbox{$D$ on wine\_white (C).}{
        \includegraphics[width=3.8cm,height=3.6cm]{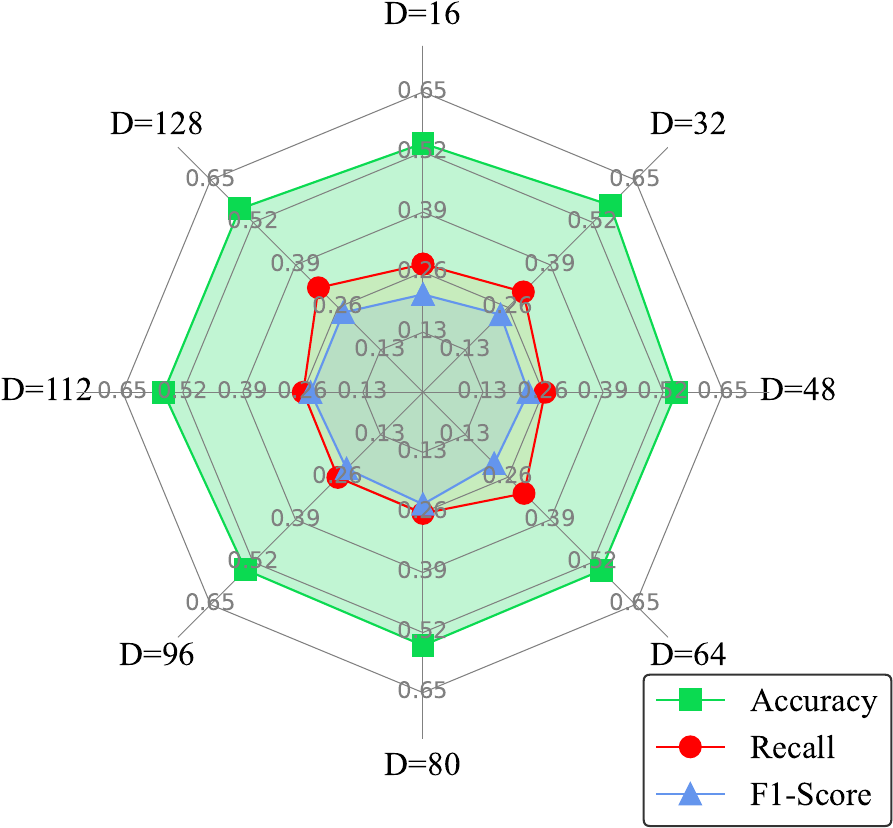}}  
    \subcaptionbox{$h$ on wine\_white (C).}{
        \includegraphics[width=3.8cm,height=3.6cm]{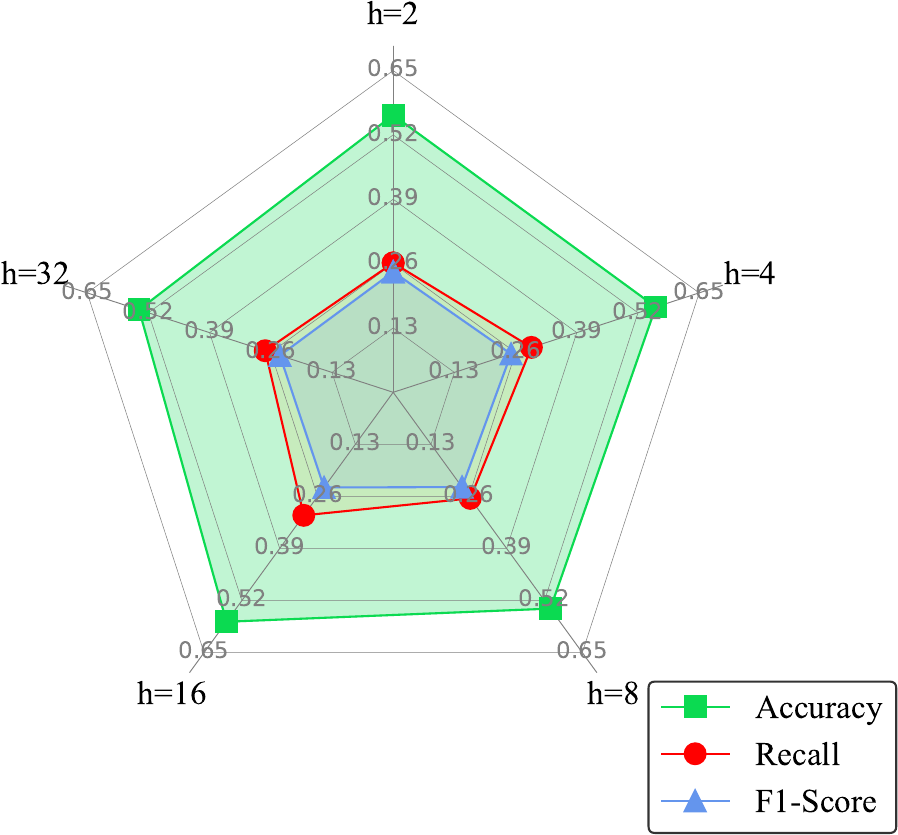}}     
    \caption{Parameter sensitivity on the number of heads $h$ and the embedding dimension $D$ on wine\_white and openml\_586.} 
    \label{fig:paras}
\end{figure}

This experiment aims to answer: \textit{How do parameters affect the performance of \model?} 
We select the wine\_white and openml\_586 datasets as representative examples. We focus on the number of heads $h$ and the embedding dimension $D$ in training procedure of \model. Specifically, We set $D=32$ and test the value of $h$ with the selected set $\{2, 4, 8, 16, 32\}$, respectively. Then we set $h=16$ and test the value of $D$ with the selected set $\{16, 32, 48, 64, 80, 96, 112, 128\}$, respectively. 
Figure~\ref{fig:paras} shows the comparison results in terms of Accuracy, Recall and F1 Score for classification tasks, and 1-MAE, 1-RMSE, $R^2$ Score for regression tasks. To better illustrate the comparison, we use 1-MAE and 1-RMSE to visualize the results so that higher metric values correspond to better performance.
In Figure~\ref{fig:paras}, we observe that the performance of downstream tasks generally remains stable across different values of $h$ and $D$, with significant changes occurring only at specific parameter values, such as $h=4$ and $D=64$ for the regression task.
A possible reason for this observation is that our proposed \model\ can effectively decouple information within feature space and can capture contextual information during evaluation process.
This observation indicates that \model\ is generally not sensitive to the number of heads $h$ and the embedding dimension $D$. 
Based on the above analysis, it can be concluded that the evaluation process of \model\ is generally robust and stable. 

\section{Conclusion}
In this paper, we propose a generalized adaptive feature space evaluator \model\ for iterative feature space optimization.
\model\ consists of two key components: feature-sample subspace generator and contextual attention evaluator.
The first component decouples the complex information within the feature space to generate diverse feature subspaces by the cooperation of the feature index optimizer and sample index optimizer.
This enhances the ability of the subsequent evaluator to capture the most challenging information for more accurate feature space evaluation.
The second component captures the intrinsic complexity of feature-sample interactions using a weight-sharing contextual-attention evaluator to ensure fair and accurate evaluation.
Considering the information overlap across consecutive iterations, we incrementally update the evaluator's parameters to retain past knowledge while incorporating new insights. This allows the evaluator to efficiently capture the evolving patterns of the feature space.
Extensive experimental results have demonstrated that \model\ has achieved superior performance compared to other baselines.
In addition, \model\ exhibits strong adaptability, generalization, and robustness in various iterative feature optimization frameworks.
In the future, we will focus on further enhancing the generalization capability of \model\ to enable it to effectively handle distribution shifts and perform robustly across different datasets.

\section{GenAI Usage Disclosure}
The authors declare that no GenAI tools were used at any stage of this research. This paper was entirely completed by the authors.

\bibliographystyle{ACM-Reference-Format}
\bibliography{reference}

@String{Computing = "Computing" }

@String{Computer = "{IEEE} Computer" }

@String{Springer = "Springer-Verlag" }

@article{zhu2023hybrid,
  title={A hybrid artificial immune optimization for high-dimensional feature selection},
  author={Zhu, Yongbin and Li, Wenshan and Li, Tao},
  journal={Knowledge-Based Systems},
  volume={260},
  pages={110111},
  year={2023},
  publisher={Elsevier}
}

@article{vommi2023hybrid,
  title={A hybrid filter-wrapper feature selection using Fuzzy KNN based on Bonferroni mean for medical datasets classification: A COVID-19 case study},
  author={Vommi, Amukta Malyada and Battula, Tirumala Krishna},
  journal={Expert Systems with Applications},
  volume={218},
  pages={119612},
  year={2023},
  publisher={Elsevier}
}

@article{htun2023survey,
  title={Survey of feature selection and extraction techniques for stock market prediction},
  author={Htun, Htet Htet and Biehl, Michael and Petkov, Nicolai},
  journal={Financial Innovation},
  volume={9},
  number={1},
  pages={26},
  year={2023},
  publisher={Springer}
}

@article{jia2022feature,
  title={Feature dimensionality reduction: a review},
  author={Jia, Weikuan and Sun, Meili and Lian, Jian and Hou, Sujuan},
  journal={Complex \& Intelligent Systems},
  volume={8},
  number={3},
  pages={2663--2693},
  year={2022},
  publisher={Springer}
}

@article{zebari2020comprehensive,
  title={A comprehensive review of dimensionality reduction techniques for feature selection and feature extraction},
  author={Zebari, Rizgar and Abdulazeez, Adnan and Zeebaree, Diyar and Zebari, Dilovan and Saeed, Jwan},
  journal={Journal of Applied Science and Technology Trends},
  volume={1},
  number={1},
  pages={56--70},
  year={2020}
}

@article{priyatno2024systematic,
  title={A Systematic Literature Review: Recursive Feature Elimination Algorithms},
  author={Priyatno, Arif Mudi and Widiyaningtyas, Triyanna and others},
  journal={JITK (Jurnal Ilmu Pengetahuan dan Teknologi Komputer)},
  volume={9},
  number={2},
  pages={196--207},
  year={2024}
}

@article{darst2018using,
  title={Using recursive feature elimination in random forest to account for correlated variables in high dimensional data},
  author={Darst, Burcu F and Malecki, Kristen C and Engelman, Corinne D},
  journal={BMC genetics},
  volume={19},
  pages={1--6},
  year={2018},
  publisher={Springer}
}

@article{liu2021automated,
  title={Automated feature selection: A reinforcement learning perspective},
  author={Liu, Kunpeng and Fu, Yanjie and Wu, Le and Li, Xiaolin and Aggarwal, Charu and Xiong, Hui},
  journal={IEEE Transactions on Knowledge and Data Engineering},
  volume={35},
  number={3},
  pages={2272--2284},
  year={2021},
  publisher={IEEE}
}

@article{wang2024mel,
  title={MEL: efficient multi-task evolutionary learning for high-dimensional feature selection},
  author={Wang, Xubin and Shangguan, Haojiong and Huang, Fengyi and Wu, Shangrui and Jia, Weijia},
  journal={IEEE Transactions on Knowledge and Data Engineering},
  year={2024},
  publisher={IEEE}
}

@article{xiao2024traceable,
  title={Traceable group-wise self-optimizing feature transformation learning: A dual optimization perspective},
  author={Xiao, Meng and Wang, Dongjie and Wu, Min and Liu, Kunpeng and Xiong, Hui and Zhou, Yuanchun and Fu, Yanjie},
  journal={ACM Transactions on Knowledge Discovery from Data},
  volume={18},
  number={4},
  pages={1--22},
  year={2024},
  publisher={ACM New York, NY}
}

@inproceedings{wang2022group,
  title={Group-wise reinforcement feature generation for optimal and explainable representation space reconstruction},
  author={Wang, Dongjie and Fu, Yanjie and Liu, Kunpeng and Li, Xiaolin and Solihin, Yan},
  booktitle={Proceedings of the 28th ACM SIGKDD Conference on Knowledge Discovery and Data Mining},
  pages={1826--1834},
  year={2022}
}

@INPROCEEDINGS{8614039,
  author={Escanilla, Nicholas Sean and Hellerstein, Lisa and Kleiman, Ross and Kuang, Zhaobin and Shull, James and Page, David},
  booktitle={2018 17th IEEE International Conference on Machine Learning and Applications (ICMLA)}, 
  title={Recursive Feature Elimination by Sensitivity Testing}, 
  year={2018},
  volume={},
  number={},
  pages={40-47},
  doi={10.1109/ICMLA.2018.00014}}

@inproceedings{huang2025time,
  title={Time-fs: joint learning of tensorial incomplete multi-view unsupervised feature selection and missing-view imputation},
  author={Huang, Yanyong and Lu, Minghui and Huang, Wei and Yi, Xiuwen and Li, Tianrui},
  booktitle={Proceedings of the AAAI Conference on Artificial Intelligence},
  volume={39},
  number={16},
  pages={17503--17510},
  year={2025}
}

@misc{openml2024,
  author = {Public},
  title = {Openml Dataset Download},
  year = {2024},
  howpublished = {[EB/OL]. \url{https://www.openml.org}},
}

@misc{uci2024,
  author = {Public},
  title = {UCI Dataset Download},
  year = {2024}, 
  howpublished = {[EB/OL]. \url{https://archive.ics.uci.edu/}}
}

@misc{kaggle2024,
  author = {Public},
  title = {Kaggle Dataset Download},
  year = {2024}, 
  howpublished = {[EB/OL]. \url{https://www.kaggle.com/c/amazon-employee-access-challenge/data}}
}

@inproceedings{hassanieh2024selective,
  title={Selective Deep Autoencoder for Unsupervised Feature Selection},
  author={Hassanieh, Wael and Chehade, Abdallah},
  booktitle={Proceedings of the AAAI Conference on Artificial Intelligence},
  volume={38},
  pages={12322--12330},
  year={2024}
}

@article{su2012linear,
  title={Linear regression},
  author={Su, Xiaogang and Yan, Xin and Tsai, Chih-Ling},
  journal={Wiley Interdisciplinary Reviews: Computational Statistics},
  volume={4},
  number={3},
  pages={275--294},
  year={2012},
  publisher={Wiley Online Library}
}

@article{kim2014predicting,
  title={Predicting restaurant financial distress using decision tree and AdaBoosted decision tree models},
  author={Kim, Soo Y and Upneja, Arun},
  journal={Economic Modelling},
  volume={36},
  pages={354--362},
  year={2014},
  publisher={Elsevier}
}

@article{li2023interpolation,
  title={Interpolation of GNSS Position Time Series Using GBDT, XGBoost, and RF Machine Learning Algorithms and Models Error Analysis},
  author={Li, Zhen and Lu, Tieding and Yu, Kegen and Wang, Jie},
  journal={Remote Sensing},
  volume={15},
  number={18},
  pages={4374},
  year={2023},
  publisher={MDPI}
}

@article{khajavi2023predicting,
  title={Predicting the carbon dioxide emission caused by road transport using a Random Forest (RF) model combined by Meta-Heuristic Algorithms},
  author={Khajavi, Hamed and Rastgoo, Amir},
  journal={Sustainable Cities and Society},
  volume={93},
  pages={104503},
  year={2023},
  publisher={Elsevier}
}

@article{nusinovici2020logistic,
  title={Logistic regression was as good as machine learning for predicting major chronic diseases},
  author={Nusinovici, Simon and Tham, Yih Chung and Yan, Marco Yu Chak and Ting, Daniel Shu Wei and Li, Jialiang and Sabanayagam, Charumathi and Wong, Tien Yin and Cheng, Ching-Yu},
  journal={Journal of clinical epidemiology},
  volume={122},
  pages={56--69},
  year={2020},
  publisher={Elsevier}
}

@article{asselman2023enhancing,
  title={Enhancing the prediction of student performance based on the machine learning XGBoost algorithm},
  author={Asselman, Amal and Khaldi, Mohamed and Aammou, Souhaib},
  journal={Interactive Learning Environments},
  volume={31},
  number={6},
  pages={3360--3379},
  year={2023},
  publisher={Taylor \& Francis}
}

@article{zhu2021class,
  title={Class-incremental learning via dual augmentation},
  author={Zhu, Fei and Cheng, Zhen and Zhang, Xu-Yao and Liu, Cheng-lin},
  journal={Advances in Neural Information Processing Systems},
  volume={34},
  pages={14306--14318},
  year={2021}
}

@article{shieh2020continual,
  title={Continual learning strategy in one-stage object detection framework based on experience replay for autonomous driving vehicle},
  author={Shieh, Jeng-Lun and Haq, Qazi Mazhar ul and Haq, Muhamad Amirul and Karam, Said and Chondro, Peter and Gao, De-Qin and Ruan, Shanq-Jang},
  journal={Sensors},
  volume={20},
  number={23},
  pages={6777},
  year={2020},
  publisher={MDPI}
}

@inproceedings{read2012batch,
  title={Batch-incremental versus instance-incremental learning in dynamic and evolving data},
  author={Read, Jesse and Bifet, Albert and Pfahringer, Bernhard and Holmes, Geoff},
  booktitle={Advances in Intelligent Data Analysis XI: 11th International Symposium, IDA 2012, Helsinki, Finland, October 25-27, 2012. Proceedings 11},
  pages={313--323},
  year={2012},
  organization={Springer}
}

@inproceedings{shim2021online,
  title={Online class-incremental continual learning with adversarial shapley value},
  author={Shim, Dongsub and Mai, Zheda and Jeong, Jihwan and Sanner, Scott and Kim, Hyunwoo and Jang, Jongseong},
  booktitle={Proceedings of the AAAI Conference on Artificial Intelligence},
  volume={35},
  pages={9630--9638},
  year={2021}
}

@article{kirkpatrick2017overcoming,
  title={Overcoming catastrophic forgetting in neural networks},
  author={Kirkpatrick, James and Pascanu, Razvan and Rabinowitz, Neil and Veness, Joel and Desjardins, Guillaume and Rusu, Andrei A and Milan, Kieran and Quan, John and Ramalho, Tiago and Grabska-Barwinska, Agnieszka and others},
  journal={Proceedings of the national academy of sciences},
  volume={114},
  number={13},
  pages={3521--3526},
  year={2017},
  publisher={National Acad Sciences}
}

@inproceedings{grosse2016kronecker,
  title={A kronecker-factored approximate fisher matrix for convolution layers},
  author={Grosse, Roger and Martens, James},
  booktitle={International Conference on Machine Learning},
  pages={573--582},
  year={2016},
  organization={PMLR}
}

@article{2021IncDet,
  title={IncDet: In Defense of Elastic Weight Consolidation for Incremental Object Detection.},
  author={ Liu, L.  and  Kuang, Z.  and  Chen, Y.  and  Xue, Jh.  and  Yang, W.  and  Zhang, W. },
  journal={IEEE transactions on neural networks and learning systems},
  volume={32},
  number={6},
  pages={2306-2319},
  year={2021},
}

@article{li2017learning,
  title={Learning without forgetting},
  author={Li, Zhizhong and Hoiem, Derek},
  journal={IEEE transactions on pattern analysis and machine intelligence},
  volume={40},
  number={12},
  pages={2935--2947},
  year={2017},
  publisher={IEEE}
}

@inproceedings{isele2018selective,
  title={Selective experience replay for lifelong learning},
  author={Isele, David and Cosgun, Akansel},
  booktitle={Proceedings of the AAAI Conference on Artificial Intelligence},
  volume={32},
  year={2018}
}

@article{shin2017continual,
  title={Continual learning with deep generative replay},
  author={Shin, Hanul and Lee, Jung Kwon and Kim, Jaehong and Kim, Jiwon},
  journal={Advances in neural information processing systems},
  volume={30},
  year={2017}
}

@article{rajasegaran2019random,
  title={Random path selection for continual learning},
  author={Rajasegaran, Jathushan and Hayat, Munawar and Khan, Salman H and Khan, Fahad Shahbaz and Shao, Ling},
  journal={Advances in neural information processing systems},
  volume={32},
  year={2019}
}

@inproceedings{serra2018overcoming,
  title={Overcoming catastrophic forgetting with hard attention to the task},
  author={Serra, Joan and Suris, Didac and Miron, Marius and Karatzoglou, Alexandros},
  booktitle={International conference on machine learning},
  pages={4548--4557},
  year={2018},
  organization={PMLR}
}

@inproceedings{aljundi2017expert,
  title={Expert gate: Lifelong learning with a network of experts},
  author={Aljundi, Rahaf and Chakravarty, Punarjay and Tuytelaars, Tinne},
  booktitle={Proceedings of the IEEE conference on computer vision and pattern recognition},
  pages={3366--3375},
  year={2017}
}

@inproceedings{messaoud2021trajectory,
  title={Trajectory prediction for autonomous driving based on multi-head attention with joint agent-map representation},
  author={Messaoud, Kaouther and Deo, Nachiket and Trivedi, Mohan M and Nashashibi, Fawzi},
  booktitle={2021 IEEE Intelligent Vehicles Symposium (IV)},
  pages={165--170},
  year={2021},
  organization={IEEE}
}

@inproceedings{sun2020generating,
  title={Generating diverse translation by manipulating multi-head attention},
  author={Sun, Zewei and Huang, Shujian and Wei, Hao-Ran and Dai, Xin-yu and Chen, Jiajun},
  booktitle={Proceedings of the AAAI Conference on Artificial Intelligence},
  volume={34},
  pages={8976--8983},
  year={2020}
}

@article{vaswani2017attention,
  title={Attention is all you need},
  author={Vaswani, A},
  journal={Advances in Neural Information Processing Systems},
  year={2017}
}

@InProceedings{Dai_2021_CVPR,
    title = {Dynamic Head: Unifying Object Detection Heads With Attentions},
    author= {Dai, Xiyang and Chen, Yinpeng and Xiao, Bin and Chen, Dongdong and Liu, Mengchen and Yuan, Lu and Zhang, Lei},
    booktitle ={Proceedings of the IEEE/CVF Conference on Computer Vision and Pattern Recognition (CVPR)},
    month= {June},
    year={2021},
    pages={7373-7382}
}

@article{guyon2002gene,
  title={Gene selection for cancer classification using support vector machines},
  author={Guyon, Isabelle and Weston, Jason and Barnhill, Stephen and Vapnik, Vladimir},
  journal={Machine learning},
  volume={46},
  pages={389--422},
  year={2002},
  publisher={Springer}
}

@inproceedings{moriya2018progressive,
  title={Progressive neural network-based knowledge transfer in acoustic models},
  author={Moriya, Takafumi and Masumura, Ryo and Asami, Taichi and Shinohara, Yusuke and Delcroix, Marc and Yamaguchi, Yoshikazu and Aono, Yushi},
  booktitle={2018 Asia-Pacific Signal and Information Processing Association Annual Summit and Conference (APSIPA ASC)},
  pages={998--1002},
  year={2018},
  organization={IEEE}
}

@article{nguyen2020survey,
  title={A survey on swarm intelligence approaches to feature selection in data mining},
  author={Nguyen, Bach Hoai and Xue, Bing and Zhang, Mengjie},
  journal={Swarm and Evolutionary Computation},
  volume={54},
  pages={100663},
  year={2020},
  publisher={Elsevier}
}

@article{pudjihartono2022review,
  title={A review of feature selection methods for machine learning-based disease risk prediction},
  author={Pudjihartono, Nicholas and Fadason, Tayaza and Kempa-Liehr, Andreas W and O'Sullivan, Justin M},
  journal={Frontiers in Bioinformatics},
  volume={2},
  pages={927312},
  year={2022},
  publisher={Frontiers Media SA}
}

@article{arora2020bolasso,
  title={A Bolasso based consistent feature selection enabled random forest classification algorithm: An application to credit risk assessment},
  author={Arora, Nisha and Kaur, Pankaj Deep},
  journal={Applied Soft Computing},
  volume={86},
  pages={105936},
  year={2020},
  publisher={Elsevier}
}

@article{nouri2021novel,
  title={A novel multi-objective forest optimization algorithm for wrapper feature selection},
  author={Nouri-Moghaddam, Babak and Ghazanfari, Mehdi and Fathian, Mohammad},
  journal={Expert Systems with Applications},
  volume={175},
  pages={114737},
  year={2021},
  publisher={Elsevier}
}

@article{liu2023novel,
  title={A novel relation aware wrapper method for feature selection},
  author={Liu, Zhaogeng and Yang, Jielong and Wang, Li and Chang, Yi},
  journal={Pattern Recognition},
  volume={140},
  pages={109566},
  year={2023},
  publisher={Elsevier}
}
\end{document}